\documentclass[10pt,journal,letterpaper]{IEEEtran}
\usepackage[T1]{fontenc}
\usepackage{cite}
\usepackage{amsmath,amssymb,amsfonts}
\usepackage{graphicx}
\usepackage{booktabs}
\usepackage{array}
\usepackage{multirow}
\usepackage[hyphens]{url}
\usepackage{xcolor}

\newcommand{\scoreup}[2]{%
    #1\rlap{%
        \hspace{0.1em}%
        {\fontsize{6}{7}\selectfont\textcolor{blue}{(#2)}}%
    }%
}

\newcommand{\scoredown}[2]{%
    #1\rlap{%
        \hspace{0.1em}%
        {\fontsize{6}{7}\selectfont\textcolor{red}{(#2)}}%
    }%
}
\title{Test-Time Selective Adaptation for Cross-Modal UAV Object Tracking with Token Shift}
\author{Yun Xiao, Hailong Ye, Zhicheng Zhao, Lei Liu, and Amir Hussain%
\thanks{Yun Xiao, Hailong Ye, and Zhicheng Zhao are with Anhui University, Hefei, China.}%
\thanks{Lei Liu is with Anhui University of Science and Technology, Huainan, China.}%
\thanks{Amir Hussain is with the SDAIA-KFUPM Joint Research Centre for AI, Information and Computer Science Department, King Fahd University of Petroleum and Minerals, Dhahran 31261, Saudi Arabia (e-mail: amir.hussain@kfupm.edu.sa).}%
}
\begin{document}
\maketitle

\begin{abstract}
Cross-modal unmanned aerial vehicle (UAV) object tracking faces modality-induced appearance discrepancies and abrupt target-position shifts that can arise from modality switching, UAV motion, and inter-sensor offsets. Although existing methods model modality and spatial variations, they primarily rely on offline representation learning and offer limited localization-oriented adaptation during testing. To address these challenges, we propose Test-Time Selective Adaptation with Token Shift (TSAT), which derives spatial and temporal adaptation signals from unlabeled test sequences for online optimization. Specifically, we design a controllable token shift auxiliary task, which imposes explicit spatial correspondence between predefined token translations and resultant localization offsets, generating stable annotation-free spatial adaptation signals. Furthermore, we introduce a temporal backward prediction auxiliary task to exploit frame-wise temporal dependencies, effectively mitigating cumulative localization errors in continuous tracking. To suppress interference from unreliable tracking samples, we develop a reliability-aware update gating strategy to screen valid adaptation signals and stabilize online parameter updates. Experiments on CM-UOT demonstrate that TSAT achieves state-of-the-art performance against existing test-time adaptation methods, exhibiting superior robustness under frequent modality switching and severe spatial displacement. Experiments across tracking frameworks and datasets further validate the effectiveness and generalization capability of TSAT.
\end{abstract}

\begin{IEEEkeywords}
Cross-modal object tracking, unmanned aerial vehicles, test-time adaptation, token shift.
\end{IEEEkeywords}

\section{Introduction}
Visual object tracking estimates a target's state in subsequent frames from its initial location in the first frame~\cite{bertinetto2016siamfc, danelljan2019atom,bhat2019dimp,yan2021stark}. Deployed on UAV platforms, it benefits from flexible mobility, adjustable aerial viewpoints, and wide-area coverage, supporting applications such as intelligent surveillance, emergency rescue, low-altitude aerial inspection, and UAV reconnaissance. However, single-modal UAV tracking is limited by modality-specific weaknesses. Visible images provide rich appearance details but are sensitive to low illumination, whereas thermal infrared images are less affected by illumination changes but lack fine-grained texture information. Multimodal UAV tracking exploits their complementarity to improve robustness in complex environments~\cite{li2019rgbtbenchmark,zhang2022vtuav}, but continuous acquisition and joint processing increase communication, computation, and power costs. To balance adaptability and efficiency, cross-modal UAV object tracking processes one modality at a time and switches between visible and thermal infrared as needed, offering a practical solution for resource-constrained UAV platforms~\cite{xiao2026sarla}.

\begin{figure}[t]
    \centering
    \includegraphics[width=\columnwidth]{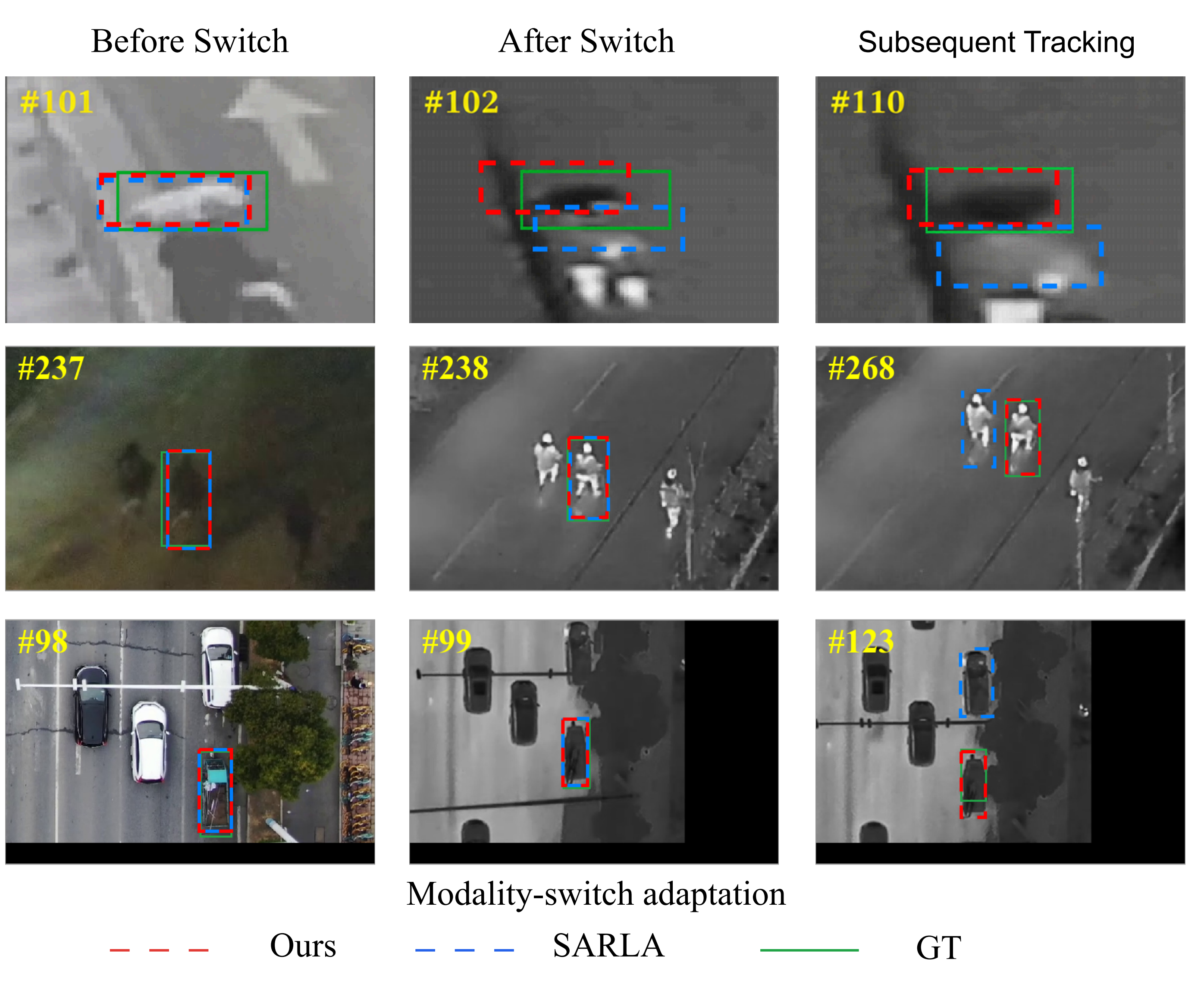}
    \caption{Tracking results before modality switching, at the switch frame, and during subsequent tracking. Red dashed, blue dashed, and green solid boxes denote TSAT, SARLA~\cite{xiao2026sarla}, and ground truth, respectively.}
    \label{fig:tracking_motivation}
\end{figure}

However, modality switching also poses two major challenges for cross-modal UAV object tracking. On the one hand, differences in imaging mechanisms and appearance distributions across modalities can lead to inconsistent feature representations during switching. On the other hand, when compounded by UAV viewpoint movement, target scale variations, and inter-sensor offsets, modality switching can further exacerbate target-position shifts between frames, making accurate localization more difficult and increasing the risk of target drift and tracking failure. Existing methods mitigate modality discrepancies through multimodal feature fusion~\cite{xiao2022apfnet, liu2024challenge}, cross-modal feature interaction~\cite{hui2023tbsi, zhu2023vipt,cao2024bat,hou2024sdstrack}, modality-adaptive representation learning, knowledge distillation, and prototype learning~\cite{li2022crossmodal, liu2024modalityfusion,li2024multiteacher,liu2025prototype}. Recent cross-modal UAV object tracking research further models modality and spatial state variations to improve robustness in dynamic aerial scenarios~\cite{xiao2026sarla}. Despite these advances, current methods remain primarily focused on offline representation learning and provide limited online correction when modality switching or target-position shifts cause localization errors. Consequently, localization errors in individual frames may propagate and accumulate through subsequent search-region initialization, eventually causing tracking drift.As illustrated in Fig.~\ref{fig:tracking_motivation}, SARLA may still localize the target at the modality-switch frame, but its localization can deteriorate during subsequent tracking as errors propagate and accumulate. These observations motivate the introduction of reliable spatial and temporal constraints during testing to stabilize localization responses and improve tracking performance after modality switching.

Test-time adaptation has emerged as an effective solution to address domain distribution shift by online fine-tuning with unlabeled test data~\cite{sun2020ttt,wang2021tent,wang2022cotta}. Different from static offline models, TTA enables continuous model adaptation to dynamic test environments, which naturally matches the iterative and time-varying characteristics of cross-modal UAV object tracking. Nevertheless, existing TTA paradigms have inherent limitations for tracking tasks. Most TTA methods are designed for classification and segmentation, optimizing models via entropy minimization, confidence estimation or parameter recovery~\cite{wang2021tent,niu2022eata,niu2023sar,shao2025pura}. They prioritize prediction certainty and parameter stability but ignore the spatial-temporal structural properties of tracking responses, failing to model localization offset correlations and suppress cumulative tracking errors.

\begin{figure}[t]
    \centering
    \includegraphics[width=\columnwidth]{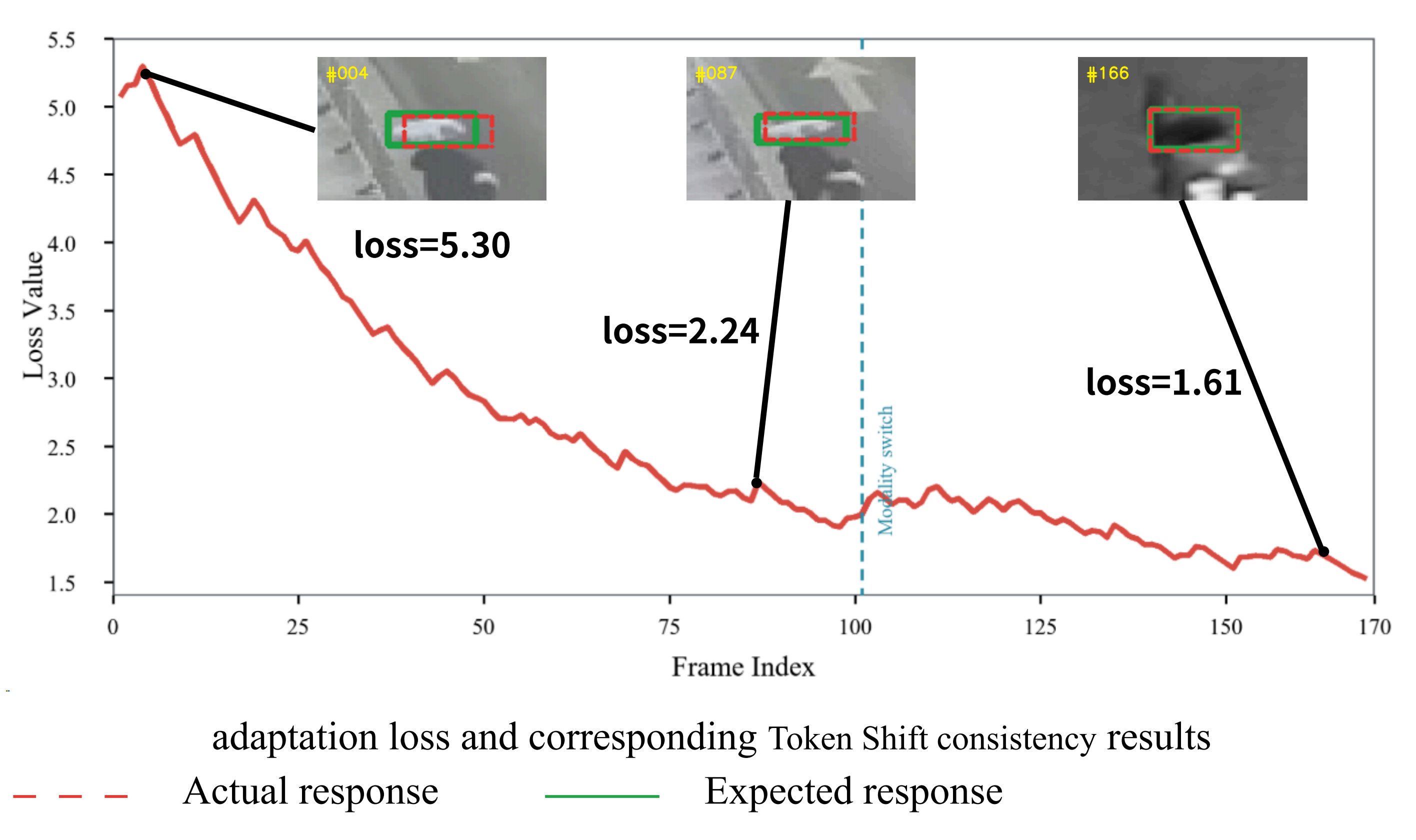}
    \caption{Adaptation loss and corresponding TSE response consistency. Red dashed and green solid boxes denote the actual and expected response locations, respectively. The vertical dashed line indicates a modality switch.}
    \label{fig:tse_motivation}
\end{figure}

To address the above gaps, we propose Test-Time Selective Adaptation with Token Shift (TSAT), a TTA framework for cross-modal UAV object tracking. Specifically, TSAT constructs two complementary self-supervised auxiliary tasks, Token Shift Equivariance (TSE) and Temporal Backward Prediction (TBP), from unlabeled test sequences to provide reliable spatial and temporal signals for online adaptation. First, TSE establishes verifiable spatial correspondence between controllable token translations and response offsets, thereby enhancing localization stability under spatial perturbations. Second, TBP uses adjacent-frame backward prediction to exploit temporal dependencies and provide a temporal constraint for online adaptation. Third, we propose Reliability-aware Update Gating (RUG), which hierarchically filters noisy samples and abnormal adaptation signals, reducing the risk of unreliable online updates amplifying localization errors during continuous tracking. As shown in Fig.~\ref{fig:tse_motivation}, the actual response becomes more closely aligned with the expected response as the adaptation loss decreases, indicating more accurate spatial adaptation. Extensive experiments on CM-UOT~\cite{xiao2026sarla} validate that TSAT achieves consistent performance gains over state-of-the-art trackers and TTA methods, especially under modality switching and large spatial displacement conditions.

The main contributions of this work are summarized as follows:

\begin{itemize}
    \item We propose Test-Time Selective Adaptation with Token Shift (TSAT), which constructs spatial and temporal self-supervised auxiliary tasks from unlabeled test sequences to address spatial response shifts and cumulative localization errors in cross-modal UAV object tracking.

    \item We design Token Shift Equivariance (TSE) as a spatial auxiliary task that derives annotation-free spatial supervision from the correspondence between controllable token shifts and localization response offsets. We further introduce a temporal constraint through the temporal auxiliary task Temporal Backward Prediction (TBP), which exploits backward prediction across adjacent frames to complement the spatial supervision and reduce error propagation.

    \item We develop Reliability-aware Update Gating (RUG), which filters unreliable samples and abnormal adaptation signals at multiple levels, enabling stable and effective online adaptation.
    
    \item Evaluations across datasets and tracking frameworks validate the effectiveness and generalization capability of TSAT. The results demonstrate performance gains over the corresponding baselines in different tracking settings, together with improved localization robustness under modality switching and spatial displacement.
\end{itemize}

\begin{figure*}[t]
    \centering
    \includegraphics[width=\textwidth]{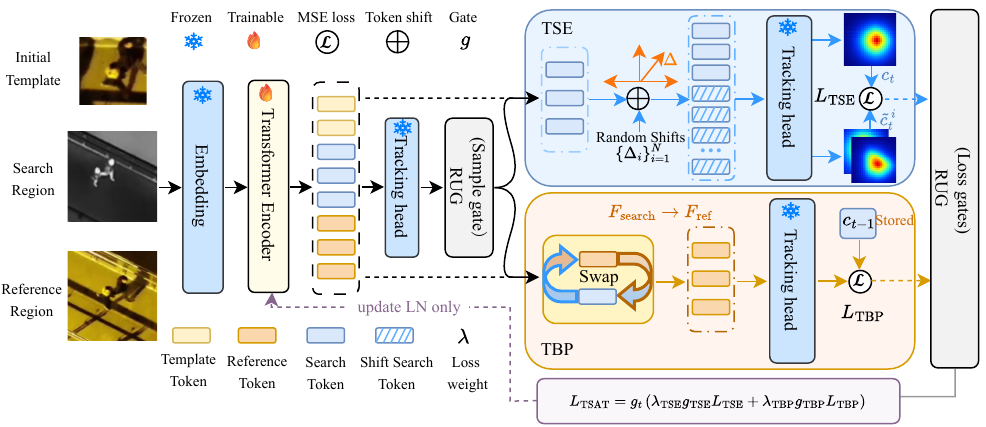}
    \caption{Overall framework of the proposed TSAT. TSAT consists of three collaborative components: Token Shift Equivariance (TSE), Temporal Backward Prediction (TBP), and Reliability-aware Update Gating (RUG).}
    \label{fig:tsat_framework}
\end{figure*}

\section{Related Work}

\subsection{Cross-modal Object Tracking}
Cross-modal object tracking exploits complementary information from different imaging modalities to improve tracking robustness in challenging scenarios, with the key objective of mitigating target-representation discrepancies caused by modality switching. Li et al.~\cite{li2022crossmodal} constructed an RGB/NIR cross-modal tracking benchmark and proposed a modality-aware dual-branch model. Subsequent studies further enhanced the adaptability of target representations to modality variations through modality-aware feature fusion~\cite{liu2024modalityfusion}, knowledge distillation~\cite{li2024multiteacher}, and prototype learning~\cite{liu2025prototype}.

Compared with fixed-view settings, cross-modal UAV object tracking is further affected by platform motion, viewpoint changes, and target-scale variations, resulting in more challenging tracking conditions and considerable practical significance. To facilitate research on this task, Xiao et al.~\cite{xiao2026sarla} established the CM-UOT benchmark and proposed state-aware representation learning to jointly model modality and spatial states, thereby improving tracking robustness in dynamic UAV scenarios. However, existing methods still largely rely on offline representation learning and focus primarily on enhancing cross-modal target representations, leaving limited capability to adapt online to dynamic spatial states during testing.

\subsection{Test-time Adaptation}

Test-time adaptation (TTA) updates a pretrained model online using unlabeled test samples to reduce the distribution mismatch between source and target domains. Wang et al.~\cite{wang2021tent} proposed TENT, which adapts normalization-layer parameters by minimizing prediction entropy. To handle continually changing test distributions, Wang et al.~\cite{wang2022cotta} proposed CoTTA, which combines prediction averaging with stochastic weight restoration to mitigate error accumulation during long-term adaptation. Niu et al.~\cite{niu2022eata} further proposed EATA to suppress noisy gradients by selecting reliable samples, while SAR~\cite{niu2023sar} integrates reliability estimation and sharpness-aware optimization to improve update stability. Recently, Shao et al.~\cite{shao2025pura} introduced TTA to RGB-T tracking, combining fast updates and parameter decomposition recovery to balance adaptation and stability.

Despite their effectiveness under distribution shifts, most existing TTA approaches are designed for classification or segmentation and primarily focus on prediction confidence and parameter stability. Although PURA extends TTA to RGB-T tracking, it mainly addresses update efficiency and parameter stability. In addition, Segu et al.~\cite{segu2023darth} propose DARTH, which uses detection consistency and patch contrastive learning to adapt object detection
and instance appearance representations in multiple object tracking at test time. Bertrand et al.~\cite{bertrand2023testtime} explore test-time training based on mask cycle consistency for matching-based video object segmentation. Cross-modal UAV object tracking also involves modality- and motion-induced spatial drift and recursively accumulated localization errors. However, these methods primarily construct adaptation objectives around prediction confidence,
parameter stability, detection consistency, or mask cycle consistency, while spatial localization changes and inter-frame localization relationships in cross-modal UAV tracking call for more targeted adaptation constraints. Compared with these methods, TSAT links its online adaptation objectives to localization responses, using TSE to constrain the correspondence between known
token translations and response-center displacements, and TBP to introduce a backward localization constraint between adjacent frames. With RUG filtering the adaptation signals, TSAT constructs an online optimization objective based on the spatial equivariance of localization responses and the localization consistency between adjacent frames, providing localization-oriented adaptation signals for cross-modal UAV tracking.

\begin{figure}[t]
    \centering
    \includegraphics[width=\columnwidth]{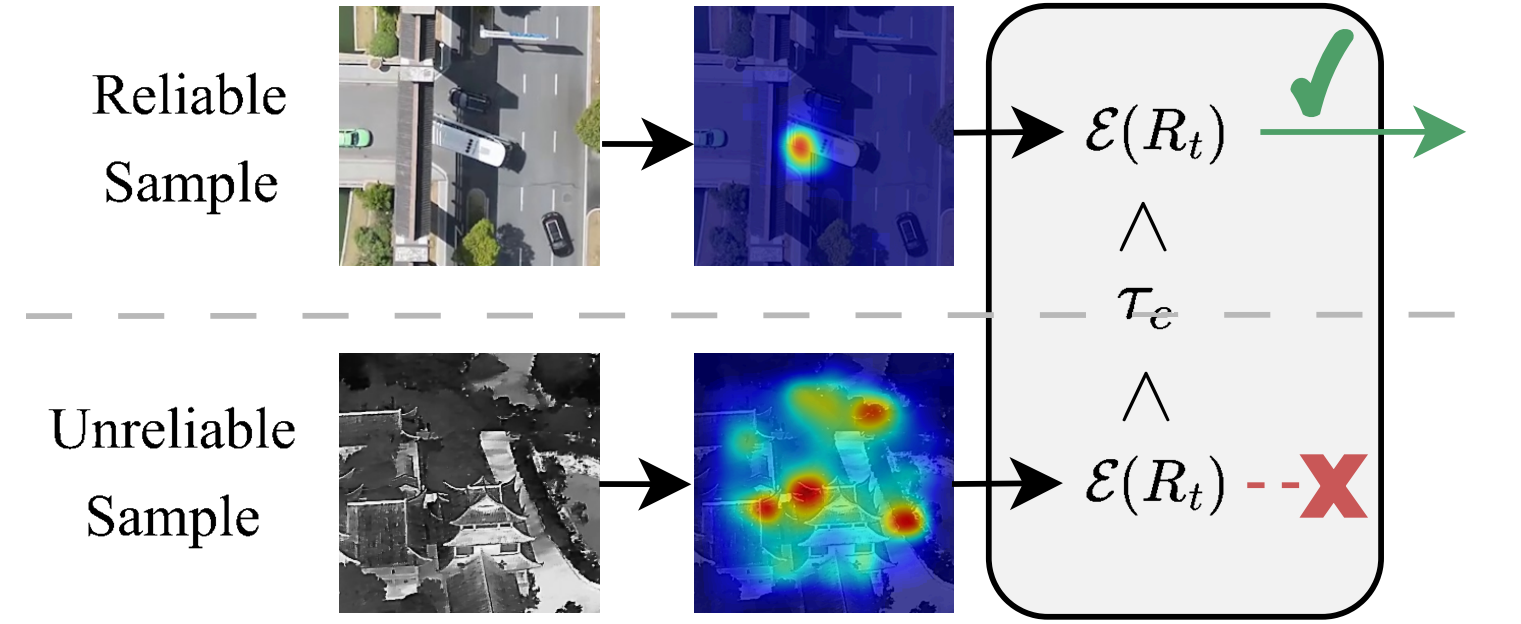}
    \caption{Illustration of entropy-based sample filtering in RUG.}
    \label{fig:rug_entropy_filtering}
\end{figure}

\section{Methodology}
This section presents Test-Time Selective Adaptation with Token Shift (TSAT), a framework for cross-modal UAV object tracking. TSAT constructs Token Shift Equivariance (TSE) and Temporal Backward Prediction (TBP) as two self-supervised auxiliary tasks from unlabeled test sequences and employs Reliability-aware Update Gating (RUG) to filter unreliable samples and abnormal adaptation signals, thereby mitigating spatial-state variations and cumulative localization errors.

\subsection{Overview}
\label{sec:overview}

We adopt SARLA~\cite{xiao2026sarla} as the baseline tracker and retain its original offline training procedure. To jointly model cross-modal appearance variations and cross-frame spatial variations, SARLA takes one initial template and two search regions as input. The initial template and current-frame search region establish cross-modal appearance correlations, while the previous-frame reference region and current-frame search region model cross-frame spatial correlations. The token features of these three inputs are processed by a Transformer encoder, after which the prediction head generates the target response map and bounding box from the current search-region features.

After obtaining the current-frame tracking result, TSAT performs online adaptation at predefined update steps, as illustrated in Fig.~\ref{fig:tsat_framework}. RUG first evaluates the reliability of the current test sample based on the entropy of its response map, as illustrated in Fig.~\ref{fig:rug_entropy_filtering}. Samples with high response entropy are regarded as unreliable and directly skip the current update. For reliable samples, TSAT constructs TSE and TBP as two self-supervised auxiliary tasks using the current test sample and historical tracking state. TSE derives spatial adaptation signals from the equivariant relationship between controllable token shifts and response offsets, while TBP derives temporal adaptation signals through backward prediction across adjacent frames to suppress cumulative localization errors. RUG further filters the adaptation signals and uses only reliable signals to update the model. The updated model subsequently processes future frames, enabling selective online adaptation to the current test sequence.

\subsection{Token Shift Equivariance Auxiliary Task}
\label{sec:TSE}

Modality switching and UAV motion may induce substantial spatial changes in the search region, causing the target response to become shifted, dispersed, or multimodal and thereby degrading localization stability. Without test annotations, directly treating the current prediction as supervision may propagate localization errors into online updates. To address this issue, we propose Token Shift Equivariance (TSE) as a localization-oriented self-supervised auxiliary task that applies controlled translations to the search-region token feature map.

TSE is based on translation equivariance. At each test-time update, we sample nonzero integer Token Shifts $\Delta_i=(d_x^i,d_y^i)$ from a predefined range. Here, $d_x^i$ and $d_y^i$ denote horizontal and vertical displacements in the token grid, respectively; positive values shift the feature map rightward and downward. We apply each shift to the current search-region token feature map by moving it by $d_x^i$ columns and $d_y^i$ rows. Tokens leaving the $H\times W$ grid are discarded, and newly exposed locations are zero-padded. Based on the translation-equivariant relationship when target features remain intact after shifting, TSE uses the predefined Token Shift as displacement supervision, constraining the response-center difference before and after shifting to match $\Delta_i$.

Given the spatially arranged search-region token feature map
$F_t \in \mathbb{R}^{C \times H \times W}$ at frame $t$, we apply a translation operator $\mathcal{T}_{\Delta}(\cdot)$ to obtain the shifted feature map $\tilde{F}_t$. The original and shifted feature maps are then fed separately into the prediction head $\mathcal{H}(\cdot)$:
\begin{equation}
\tilde{F}_t = \mathcal{T}_{\Delta}(F_t), \qquad
R_t = \mathcal{H}(F_t), \qquad
\tilde{R}_t = \mathcal{H}(\tilde{F}_t),
\label{eq:TSE_response}
\end{equation}
where $\Delta=(d_x,d_y)$ denotes the sampled Token Shift, while $R_t$ and $\tilde{R}_t$ are the response maps before and after the shift, respectively. To obtain differentiable target locations, we compute the response centers using SoftArgmax:
\begin{equation}
\mathbf{c}_t
=
\operatorname{SoftArgmax}(R_t), \qquad
\tilde{\mathbf{c}}_t
=
\operatorname{SoftArgmax}(\tilde{R}_t).
\label{eq:TSE_center}
\end{equation}

Specifically, we apply the same spatial weighting to the original and shifted response maps and denote the response used for center estimation by $\bar{R}$. The response values are then converted into normalized weights using a spatial softmax with a temperature parameter, and the response center is computed as the weighted average of spatial coordinates:
\[
\begin{aligned}
p_j(R)
&=
\frac{\exp(\bar{R}_j/\kappa)}
{\sum_{k=1}^{HW}\exp(\bar{R}_k/\kappa)},\\
\operatorname{SoftArgmax}(R)
&=
\sum_{j=1}^{HW}p_j(R)q_j.
\end{aligned}
\]
Here, $\kappa>0$ is the temperature parameter controlling the concentration of the response distribution, and $q_j$ denotes the two-dimensional coordinate of the $j$-th spatial location relative to the grid center. The centers and their displacements are expressed in token-grid units. The original and shifted response maps follow the same center estimation procedure, allowing their center difference to be compared with the sampled Token Shift. This differentiable computation connects the response map to the localization constraint, enabling the shifted response
to provide gradient signals for test-time parameter updates.

Instead of constraining the absolute location of the current prediction, TSE focuses on the relative localization change induced by Token Shift. At each test-time update, we construct $N$ Token Shifts and align the resulting response-center displacements with the corresponding sampled offsets:
\begin{equation}
\mathcal{L}_{\mathrm{TSE}}
=
\frac{1}{N}
\sum_{i=1}^{N}
\operatorname{MSE}
\left(
\tilde{\mathbf{c}}_t^{\,i}-\mathbf{c}_t,
\Delta_i
\right),
\label{eq:TSE_loss}
\end{equation}
where $\Delta_i$ denotes the $i$-th sampled Token Shift, and $\tilde{\mathbf{c}}_t^{\,i}$ is the response center obtained under the corresponding shift. The response-center displacement and Token Shift are represented in the same token-grid coordinate system. Since the supervision is directly derived from the sampled shifts, TSE requires neither manual annotations nor predicted bounding boxes as pseudo-labels, allowing spatial adaptation signals to be constructed directly from unlabeled test samples.

\subsection{Temporal Backward Prediction Auxiliary Task}
\label{sec:TBP}

TSE constructs spatial adaptation signals from the localization changes induced by Token Shift within a single frame. However, UAV object tracking is a recursive process in which the localization result of the current frame is used to determine the search region for subsequent frames. Once the predicted location deviates from the target, the resulting localization error may gradually accumulate throughout the tracking sequence. To exploit the temporal relationship between adjacent frames, we propose Temporal Backward Prediction (TBP) as a temporal self-supervised auxiliary task, which converts the stored historical tracking state into an adaptation signal.

When processing frame $t$, the tracker retains the target state from the previous frame and obtains the current-frame search region and previous-frame reference region. Together with the initial template, these two regions are processed by the shared encoder to produce feature representations containing cross-frame context. The current-frame search features are used for standard forward localization, while the previous-frame reference features are used to construct the backward prediction task.

TBP switches the prediction region from the current-frame search region to the previous-frame
reference region and reuses the existing tracking head to generate the backward response. Since
the previous-frame reference features incorporate current-frame information through joint encoding,
this response exploits cross-frame context to estimate the target location in the historical
reference region. Forward localization and backward prediction share the same prediction-head parameters, allowing the temporal constraint to act on the existing tracking representations without introducing an additional learnable prediction network.

Each spatial location in the backward response map corresponds to a position in the previous-frame
reference region. TBP therefore imposes its localization constraint on the historical reference
region, while TSE constructs a spatial constraint from controlled shifts of the current-frame features. The backward prediction produces a response map $R_{t\rightarrow t-1}$, from which the response center is obtained using SoftArgmax
\begin{equation}
\mathbf{c}_{t\rightarrow t-1}
=
\operatorname{SoftArgmax}
\left(
R_{t\rightarrow t-1}
\right).
\label{eq:TBP_center}
\end{equation}

Since the previous-frame reference region is centered on the stored target state, the backward-predicted response center should align with the previous-frame target center $\mathbf{c}_{t-1}$, which corresponds to zero displacement in the local coordinate system of the reference region. Accordingly, the TBP loss is defined as
\begin{equation}
\mathcal{L}_{\mathrm{TBP}}
=
\operatorname{MSE}
\left(
\mathbf{c}_{t\rightarrow t-1},
\mathbf{c}_{t-1}
\right).
\label{eq:TBP_loss}
\end{equation}

TBP uses stored tracking state to impose an adjacent-frame temporal constraint. Together with TSE's intra-frame Token Shift constraint, it provides complementary spatial-temporal signals from unlabeled test sequences.

\subsection{Reliability-Aware Update Gating}
\label{sec:rug}

In unlabeled test sequences, unreliable predictions can produce erroneous signals and destabilize updates. In particular, modality switching and interference from similar backgrounds may lead to dispersed or multimodal target responses. To prevent such abnormal responses from participating in model updates, we design Reliability-aware Update Gating (RUG), which filters test samples and their adaptation signals at both the response and branch-loss levels.

First, RUG employs response entropy to assess the reliability of the current test sample. Using the normalized response weights $p_j(R)$ defined for SoftArgmax, the response entropy is computed as
\[
E(R)=
-\frac{1}{\log(HW)}
\sum_{j=1}^{HW}p_j(R)\log p_j(R).
\]
Here, $HW$ denotes the number of spatial locations in the response map. Dividing by the maximum entropy $\log(HW)$ normalizes this measure to $[0,1]$. The entropy is low when the response is concentrated at a small number of locations and approaches one when the response becomes uniform across the grid.

The response center and entropy capture different properties of the same spatial distribution:
the former describes location, whereas the latter describes concentration. RUG therefore uses entropy to assess the spatial uncertainty associated with the localization response.
As illustrated in Fig.~\ref{fig:rug_entropy_filtering}, a concentrated response map with a clear target peak generally has low entropy, whereas a dispersed response has higher entropy, indicating greater localization uncertainty. The response entropy gate is defined as
\begin{equation}
g_t
=
\mathbb{I}
\left[
\mathcal{E}(R_t)<\tau_e
\right],
\label{eq:rug_entropy_gate}
\end{equation}
where $E(R_t)$ denotes the entropy of the current response map $R_t$, and $\tau_e$ is the response entropy threshold. When $g_t=1$, the current sample is allowed to participate in subsequent adaptation. When $g_t=0$, the current update is skipped. For frames passing the sample-level gate,
RUG further applies the same entropy criterion to the response maps generated by individual Token Shifts in TSE. Since different shifts may produce different response distributions, each shifted response is assessed independently. The set of accepted shift indices is defined as
\[
\mathcal{A}_t=
\left\{
i\in\{1,\ldots,N\}
\mid E(\tilde{R}_t^i)<\tau_e
\right\}.
\]
When $\mathcal{A}_t$ is nonempty, TSE aggregates only the displacement losses associated with the accepted shifts:
\[
\widehat{L}_{\mathrm{TSE}}^t
=
\frac{1}{|\mathcal{A}_t|}
\sum_{i\in\mathcal{A}_t}
\operatorname{MSE}
\left(
\tilde{c}_t^i-c_t,\Delta_i
\right).
\]
Here, $|\mathcal{A}_t|$ denotes the number of shifts participating in loss computation. Averaging over the accepted shifts keeps the aggregate loss on the scale of the mean per-shift loss. When all shifts are accepted, this expression reduces to Eq.~(3). The selected loss $\widehat{L}_{\mathrm{TSE}}^t$ serves as $L_{\mathrm{TSE}}^t$ in the branch-loss gate and overall
objective below. If no shift is accepted, the TSE branch does not contribute to the current update, while the TBP branch may still contribute after passing its own branch-loss gate.

Based on the response entropy gate, RUG further evaluates the reliability of the adaptation signals produced by TSE and TBP. Even when the current sample passes the entropy gate, an individual branch may still produce an abnormal loss. For branch $m\in\{\mathrm{TSE},\mathrm{TBP}\}$, the branch-loss gate is defined as
\begin{equation}
\begin{aligned}
g_m^t
&=
\mathbb{I}\!\left[
\mathcal{L}_m^t < \rho_m \tau_m^t
\right], \\
\tau_m^t
&=
\alpha \tau_m^{t-1}
+
(1-\alpha)\mathcal{L}_m^{t-1}.
\end{aligned}
\label{eq:rug_loss_gate}
\end{equation}
where $\mathcal{L}_m^t$ denotes the loss of branch $m$ at frame $t$, $\tau_m^t$ is the corresponding dynamic loss threshold, $\alpha$ is the exponential moving average coefficient, and $\rho_m$ is a branch-specific threshold weight controlling the tolerance of the loss gate. When $g_m^t=0$, the corresponding branch is excluded from the current parameter update. Through this hierarchical filtering strategy, RUG first rejects test samples with unreliable response states and then filters adaptation signals with abnormal losses. Consequently, TSAT performs selective updates using only reliable information, reducing the adverse effects of abnormal responses on online adaptation.

\subsection{Overall Objective and Test-Time Update}
\label{sec:online_update}

Based on the above design, TSAT integrates the TSE and TBP auxiliary tasks into a unified test-time optimization objective. The overall loss is formulated as:
\begin{equation}
\begin{aligned}
\mathcal{L}_{TSAT}^{t}
&=
g_t\Big(
\lambda_{TSE}g_{TSE}^{t}\mathcal{L}_{TSE}^{t}
\\
&\qquad+
\lambda_{TBP}g_{TBP}^{t}\mathcal{L}_{TBP}^{t}
\Big).
\end{aligned}
\label{eq:TSAT_loss}
\end{equation}

Here, $\mathcal{L}_{TSE}^{t}$ and $\mathcal{L}_{TBP}^{t}$ are the TSE and TBP losses at frame $t$, respectively, and $\lambda_{TSE}$ and $\lambda_{TBP}$ are the corresponding weights. $g_t$ is the sample-level response entropy gate, while $g_{TSE}^{t}$ and $g_{TBP}^{t}$ are the corresponding branch-loss gates, which jointly select adaptation signals for online updates.

During test-time updates, TSAT optimizes only the scale and bias parameters of LayerNorm~\cite{ba2016layernorm} layers in backbone Transformer blocks~\cite{dosovitskiy2021vit}.
These parameters operate on normalized features, allowing the model to adapt to the current test sequence by adjusting feature-channel magnitudes and offsets. The parameters of the attention layers, feed-forward networks, and prediction head remain fixed. Gradients from the auxiliary-task losses propagate through the tracking network and update only the selected LayerNorm parameters. This restricted update preserves pretrained representations and reduces overfitting during unlabeled adaptation. By combining reliable auxiliary-task signals with lightweight parameter updates, TSAT achieves stable adaptation on test sequences and further improves the robustness of cross-modal UAV object tracking.

\section{Experiments and Discussion}

\subsection{Dataset and Evaluation Metrics}
\label{sec:dataset_metrics}

\noindent\textbf{Dataset.}
We evaluate the proposed TSAT on CM-UOT~\cite{xiao2026sarla}, a benchmark for cross-modal UAV object tracking. CM-UOT contains 1,079 cross-modal sequences (>671K frames), including 863 training and 216 test sequences. CM-UOT covers diverse UAV scenarios and modality and spatial variations, enabling evaluation under modality switching and dynamic spatial changes.

\noindent\textbf{Evaluation Metrics.}
Following prior work, we adopt the one-pass evaluation (OPE) protocol and report Precision Rate (PR), Normalized Precision Rate (NPR), and Success Rate (SR). PR measures the percentage of frames whose center localization error is below 20 pixels. NPR normalizes center error by target size to reduce scale effects. SR is the area under the success plot based on the IoU between predicted and ground-truth boxes.

\begin{table}[t]
\centering
\small
\caption{Quantitative comparison with representative trackers retrained on CM-UOT. TSAT is applied to the SARLA baseline.}
\label{tab:cmuot_comparison}
\renewcommand{\arraystretch}{1.08}
\setlength{\tabcolsep}{5pt}

\begin{tabular}{l|c|ccc}
\toprule
\textbf{Method} & \textbf{Pub. Info.}
& \textbf{PR} & \textbf{NPR} & \textbf{SR} \\
\midrule
TransT~\cite{chen2021transt}
& CVPR'21 & 39.5 & 33.6 & 29.2 \\
OSTrack~\cite{ye2022ostrack}
& ECCV'22 & 37.0 & 32.7 & 28.0 \\
HiT~\cite{kang2023hit}
& ICCV'23 & 40.0 & 34.3 & 30.1 \\
SeqTrack~\cite{chen2023seqtrack}
& CVPR'23 & 43.3 & 39.2 & 33.5 \\
GRM~\cite{gao2023grm}
& CVPR'23 & 38.5 & 33.5 & 28.8 \\
EVPTrack~\cite{shi2024evptrack}
& AAAI'24 & 46.5 & 41.7 & 35.7 \\
AVTrack~\cite{li2024avtrack}
& ICML'24 & 38.0 & 32.0 & 28.1 \\
AQATrack~\cite{xie2024aqatrack}
& CVPR'24 & 44.5 & 40.1 & 34.2 \\
SUTrack~\cite{chen2025sutrack}
& AAAI'25 & 48.0 & 42.3 & 36.7 \\
SpikeTrack~\cite{zhang2026spiketrack}
& CVPR'26 & 45.1 & 40.5 & 34.8 \\
SARLA~\cite{xiao2026sarla}
& TCSVT'26 & 50.6 & 44.8 & 38.0 \\
\midrule
Ours & --
& \textbf{53.19} & \textbf{47.28} & \textbf{40.23} \\
\bottomrule
\end{tabular}
\end{table}

\begin{table*}[!t]
\centering
\small
\caption{Comparison with representative test-time adaptation
methods on the CM-UOT dataset before and after modality switching
and over the entire test sequences.}
\label{tab:first_switch}

\renewcommand{\arraystretch}{1.12}
\setlength{\tabcolsep}{2pt}

\begin{tabular*}{\textwidth}{
    @{\extracolsep{\fill}}
    l
    c
    *{9}{>{\centering\arraybackslash}p{0.062\textwidth}}
    @{\hspace{14pt}}
}
\toprule
\multirow{2}{*}{\textbf{Method}}
& \multirow{2}{*}{\textbf{Pub. Info.}}
& \multicolumn{3}{c}{\textbf{PR}}
& \multicolumn{3}{c}{\textbf{NPR}}
& \multicolumn{3}{c}{\textbf{SR}} \\
\cmidrule(lr){3-5}
\cmidrule(lr){6-8}
\cmidrule(lr){9-11}
&
& \textbf{Before} & \textbf{After} & \textbf{All}
& \textbf{Before} & \textbf{After} & \textbf{All}
& \textbf{Before} & \textbf{After} & \textbf{All} \\
\midrule

AdaBN~\cite{li2018adabn} & PR'18
& 90.99 & 43.12 & 50.42
& 86.48 & 37.32 & 44.69
& 71.50 & 32.16 & 38.05 \\

TENT~\cite{wang2021tent} & ICLR'21
& 90.97 & 42.46 & 49.79
& 86.71 & 36.54 & 43.98
& 71.67 & 31.65 & 37.57 \\

CoTTA~\cite{wang2022cotta} & CVPR'22
& 90.97 & 40.81 & 48.55
& 86.62 & 35.32 & 43.08
& 71.66 & 30.57 & 36.77 \\

ETA/EATA~\cite{niu2022eata} & ICML'22
& 90.90 & 42.30 & 49.55
& 86.60 & 36.56 & 43.89
& 71.61 & 31.57 & 37.42 \\

MEMO~\cite{zhang2022memo} & NeurIPS'22
& 91.18 & 43.89 & 51.19
& 86.68 & 37.86 & 45.24
& 71.71 & 32.72 & 38.61 \\

SAR~\cite{niu2023sar} & ICLR'23
& 90.78 & 41.87 & 49.27
& 86.51 & 36.18 & 43.68
& 71.58 & 31.36 & 37.32 \\

RoTTA~\cite{yuan2023rotta} & CVPR'23
& 91.07 & 42.23 & 49.74
& 86.37 & 37.16 & 44.53
& 71.41 & 31.92 & 37.83 \\

DeYO~\cite{lee2024deyo} & ICLR'24
& 90.88 & 42.83 & 50.06
& 86.62 & 37.24 & 44.50
& 71.58 & 32.14 & 37.92 \\

SiamABC~\cite{zaveri2025siamabc} & WACV'25
& \textbf{91.27} & 43.32 & 50.90
& 86.66 & 37.64 & 45.20
& 71.67 & 32.41 & 38.44 \\

PURA~\cite{shao2025pura} & CVPR'25
& 90.73 & 43.00 & 50.27
& 86.32 & 37.27 & 44.62
& 71.36 & 32.10 & 37.98 \\

GOLD~\cite{lai2026gold} & CVPR'26
& 91.25 & 43.03 & 50.68
& 86.79 & 37.46 & 45.11
& 71.71 & 32.32 & 38.42 \\

CAZO~\cite{zhang2026curvature} & CVPR'26
& \textbf{91.27} & 43.47 & 50.97
& \textbf{86.82} & 37.88 & 45.39
& 71.72 & 32.66 & 38.64 \\

IMSE~\cite{baek2026imse} & ICLR'26
& 91.12 & 42.82 & 50.23
& 86.75 & 37.36 & 44.81
& \textbf{71.85} & 32.25 & 38.21 \\

\midrule

SARLA (Baseline) & TCSVT'26
& 91.10 & 43.30 & 50.60
& 86.50 & 37.50 & 44.80
& 71.70 & 32.20 & 38.00 \\

Ours & --
& \scoreup{91.17}{+0.07}
& \scoreup{\textbf{46.31}}{+3.01}
& \scoreup{\textbf{53.19}}{+2.59}
& \scoreup{86.67}{+0.17}
& \scoreup{\textbf{40.36}}{+2.86}
& \scoreup{\textbf{47.28}}{+2.48}
& \scoredown{71.67}{-0.03}
& \scoreup{\textbf{34.69}}{+2.49}
& \scoreup{\textbf{40.23}}{+2.23} \\

\bottomrule
\end{tabular*}
\end{table*}

\begin{table*}[t]
\centering
\fontsize{8}{9.5}\selectfont
\caption{Comparison of representative TTA methods across ground-truth target-center displacement ranges on CM-UOT. Displacements are measured in pixels between consecutive frames; scores are evaluated over the subsequent 30 frames. The best and second-best results are highlighted in bold and blue, respectively.}
\label{tab:spatial_shift}
\renewcommand{\arraystretch}{1.10}
\setlength{\tabcolsep}{3pt}

\begin{tabular*}{\textwidth}
{@{\extracolsep{\fill}}l*{15}{c}@{}}
\toprule
\multirow[c]{2}{*}{\textbf{Method}}
& \multicolumn{3}{c}{\textbf{20--40}}
& \multicolumn{3}{c}{\textbf{40--60}}
& \multicolumn{3}{c}{\textbf{60--80}}
& \multicolumn{3}{c}{\textbf{80--100}}
& \multicolumn{3}{c}{\textbf{$>$100}} \\
\cmidrule(lr){2-4}
\cmidrule(lr){5-7}
\cmidrule(lr){8-10}
\cmidrule(lr){11-13}
\cmidrule(lr){14-16}
& \textbf{PR} & \textbf{NPR} & \textbf{SR}
& \textbf{PR} & \textbf{NPR} & \textbf{SR}
& \textbf{PR} & \textbf{NPR} & \textbf{SR}
& \textbf{PR} & \textbf{NPR} & \textbf{SR}
& \textbf{PR} & \textbf{NPR} & \textbf{SR} \\
\midrule
AdaBN
& 49.90 & 44.40 & 37.77
& 46.56 & 41.45 & 35.08
& 48.62 & 41.88 & 34.24
& 35.41 & 33.82 & 26.21
& \textcolor{blue}{33.84} & 30.74 & 25.96 \\

TENT
& 48.20 & 42.51 & 36.39
& 46.06 & 40.40 & 34.96
& 47.29 & 42.00 & 33.88
& \textcolor{blue}{46.57} & 42.81 & 33.70
& 29.83 & 27.47 & 23.17 \\

CoTTA
& 50.81 & 44.89 & 38.30
& 43.49 & 39.23 & 33.30
& 46.95 & 42.15 & 33.82
& \textbf{47.30} & \textbf{44.86} & \textbf{35.44}
& 27.24 & 24.44 & 21.06 \\

ETA/EATA
& 48.86 & 43.64 & 36.94
& 45.32 & 40.05 & 34.23
& 46.83 & 40.99 & 33.09
& 43.09 & 41.15 & 31.88
& 30.22 & 27.85 & 23.38 \\

MEMO
& 49.73 & 44.13 & 37.47
& 48.91 & 43.43 & 36.78
& \textcolor{blue}{49.74}
& \textcolor{blue}{42.42}
& \textcolor{blue}{34.66}
& 41.38 & 39.87 & 31.40
& 33.39 & 30.78 & 26.16 \\

SAR
& 49.74 & 43.96 & 37.39
& 45.27 & 39.85 & 34.17
& 45.49 & 40.95 & 32.92
& 42.73 & 40.45 & 32.11
& 28.97 & 26.67 & 22.67 \\

RoTTA
& 49.55 & 43.44 & 37.13
& \textcolor{blue}{49.80}
& \textcolor{blue}{44.97}
& \textcolor{blue}{38.20}
& 41.18 & 36.80 & 29.45
& 41.28 & 39.58 & 30.57
& 32.07 & 29.76 & 24.91 \\

DeYO
& 49.74 & 44.36 & 37.52
& 44.12 & 38.72 & 33.31
& 46.34 & 41.13 & 33.06
& 43.23 & 42.49 & 33.18
& 31.00 & 28.09 & 23.82 \\

PURA
& 49.73 & 43.86 & 37.37
& 48.59 & 43.40 & 36.83
& 48.13 & 41.45 & 33.85
& 37.09 & 35.74 & 27.83
& 32.58 & 30.46 & 25.72 \\

CAZO
& 50.70 & 44.84 & 38.24
& 48.76 & 43.48 & 36.80
& 46.91 & 41.27 & 33.63
& 39.47 & 38.12 & 29.73
& 33.25
& \textcolor{blue}{31.00}
& \textcolor{blue}{26.38} \\

GOLD
& \textcolor{blue}{51.87}
& \textcolor{blue}{46.01}
& \textcolor{blue}{39.20}
& 48.38 & 43.12 & 36.53
& 47.49 & 41.96 & 34.18
& 39.59 & 38.16 & 29.74
& 32.77 & 30.55 & 26.02 \\

IMSE
& 48.93 & 44.20 & 37.65
& 45.30 & 40.89 & 34.52
& 47.92 & 41.78 & 34.09
& 42.58 & 40.06 & 31.44
& 31.73 & 28.89 & 24.63 \\

\midrule
SARLA (Baseline)
& 49.71 & 43.57 & 37.35
& 48.62 & 42.43 & 36.26
& 40.22 & 35.38 & 28.98
& 42.21 & 40.08 & 31.27
& 27.72 & 24.83 & 20.84 \\

Ours
& \textbf{52.58} & \textbf{46.28} & \textbf{39.65}
& \textbf{51.33} & \textbf{46.09} & \textbf{38.90}
& \textbf{50.37} & \textbf{43.92} & \textbf{36.03}
& 45.76 & \textcolor{blue}{44.43} & \textcolor{blue}{34.29}
& \textbf{35.39} & \textbf{32.51} & \textbf{27.41} \\
\bottomrule
\end{tabular*}
\end{table*}

\begin{figure*}[t]
    \centering
    \includegraphics[width=\textwidth]{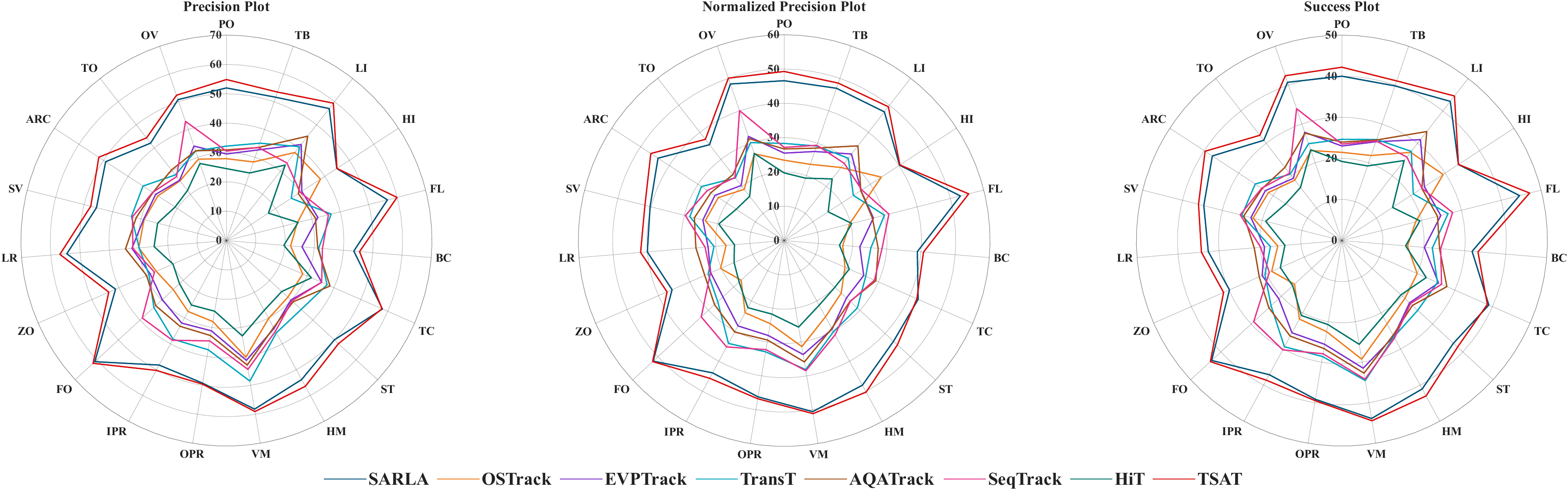}
    \caption{Attribute-based performance comparison with representative trackers on CM-UOT. From left to right, the radar charts report PR, NPR, and SR over 19 challenge attributes.}
    \label{fig:attribute_radar_trackers}
\end{figure*}

\begin{figure}[t]
    \centering
    \includegraphics[width=\columnwidth]{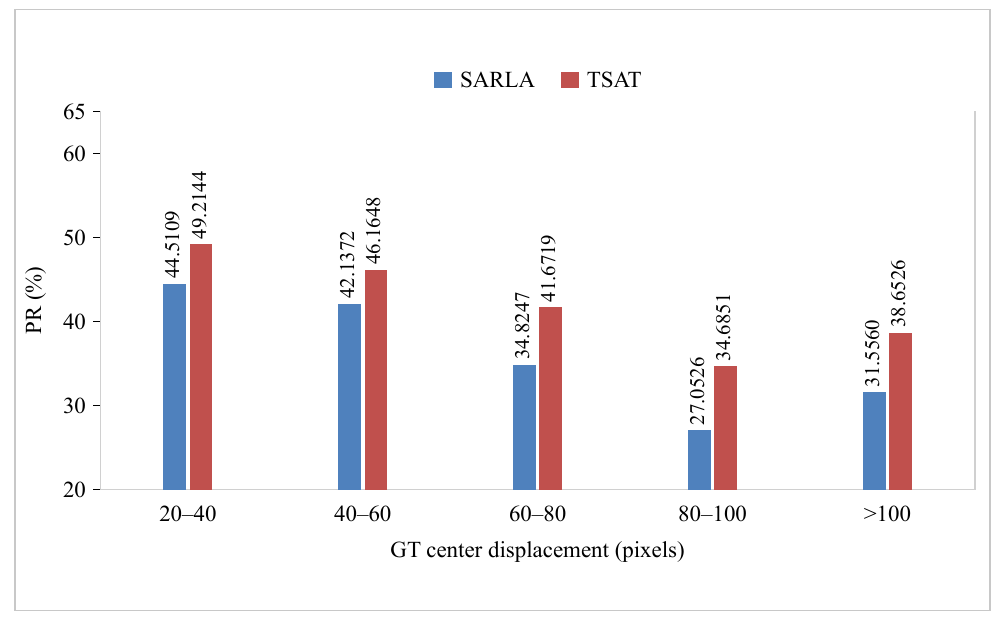}
    \caption{PR comparison after different spatial
    displacements on CM-UOT, evaluated from the first
    corresponding displacement frame to the sequence end.
    Blue and red bars denote SARLA and TSAT, respectively.}
    \label{fig:displacement_pr}
\end{figure}

\subsection{Implementation Details}
\label{sec:implementation}

We adopt SARLA~\cite{xiao2026sarla} as the baseline tracker and introduce TSAT at test time without modifying its offline training procedure. The model employs ViT-B/16~\cite{dosovitskiy2021vit} as the backbone and follows the original training configuration of SARLA. During testing, most pretrained parameters are frozen, and only the affine parameters of the LayerNorm~\cite{ba2016layernorm} layers within the Transformer blocks are updated.

TSAT performs a test-time update every 10 frames with a learning rate of $6\times10^{-5}$, and three Token Shifts are constructed for each update. The response entropy threshold $\tau_e$ in Eq.~\eqref{eq:rug_entropy_gate} is set to 0.24. The loss weights $\lambda_{\mathrm{TSE}}$ and $\lambda_{\mathrm{TBP}}$ in Eq.~\eqref{eq:TSAT_loss} are set to 220 and 50, respectively. For the compared methods, we use the default hyperparameters reported in their original papers or official implementations. All experiments are conducted on a single NVIDIA RTX 4090 GPU.

\subsection{Evaluation on CM-UOT}
\label{sec:cmuot_evaluation}

\subsubsection{Comparison with Representative Trackers}
To evaluate the overall tracking performance of TSAT, we compare it with several representative trackers on the CM-UOT test set, as reported in Table~\ref{tab:cmuot_comparison}. All trackers are retrained on the CM-UOT training set. Among the existing trackers, SARLA achieves the strongest performance, with PR, NPR, and SR scores of 50.60\%, 44.80\%, and 38.00\%, respectively.

After introducing TSAT into the SARLA baseline, the PR, NPR, and SR scores increase to 53.19\%, 47.28\%, and 40.23\%, respectively, achieving the best results across all three metrics. Compared with SARLA, TSAT improves PR, NPR, and SR by 2.59, 2.48, and 2.23 percentage points, respectively. Compared with SUTrack, the corresponding improvements are 5.19, 4.98, and 3.53 percentage points. These results demonstrate that selective adaptation using unlabeled test sequences can further improve the baseline tracker, validating the effectiveness of TSAT for cross-modal UAV object tracking.

\subsubsection{Qualitative Comparison}
\begin{figure*}[t]
    \centering
    \includegraphics[width=0.95\textwidth]{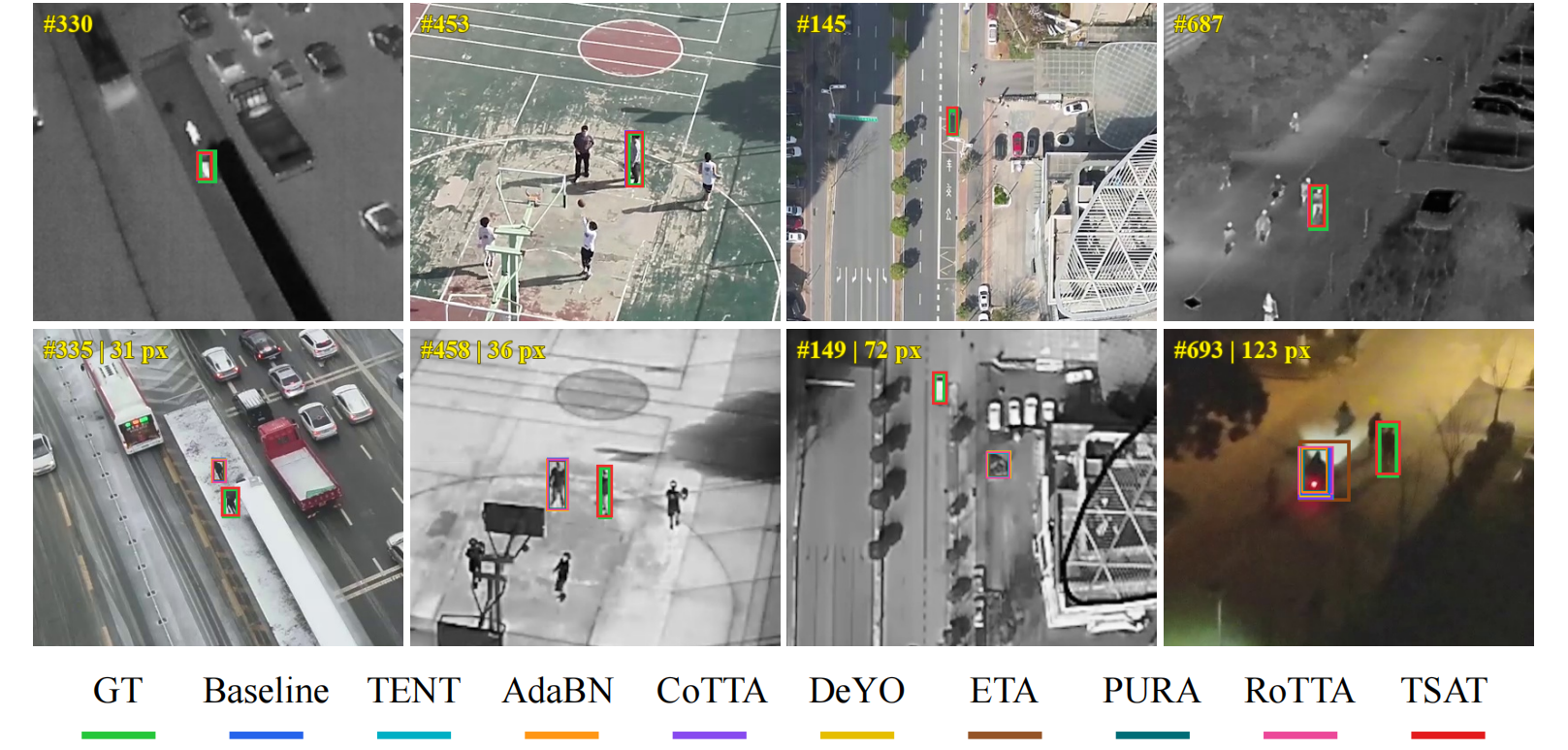}
    \caption{Qualitative comparison of different test-time adaptation methods before and after the modality switch. The displacement values in the second row indicate the ground-truth target-center displacement between the current frame and its preceding frame.}
    \label{fig:tta_qualitative}
\end{figure*}

\begin{figure}[t]
    \centering
    \includegraphics[width=\columnwidth]{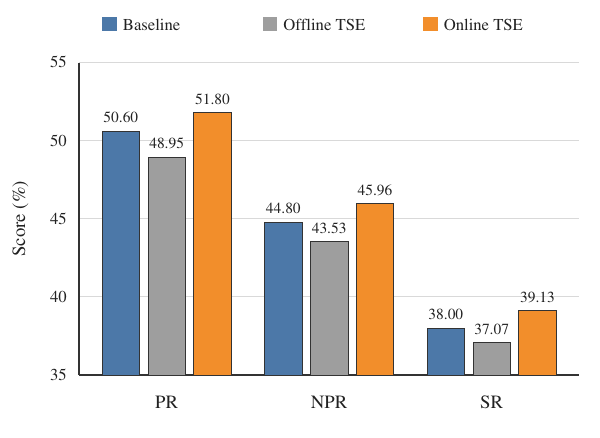}
    \caption{Comparison of Offline TSE and Online TSE on CM-UOT. The vertical axis ranges from 35\% to 55\%.}
    \label{fig:tse_stage_comparison}
\end{figure}

\subsubsection{Comparison with Test-Time Adaptation Methods}
To evaluate the test-time adaptation capability of TSAT, we compare it with several representative TTA methods. For fair comparison, we implement each method on SARLA by replacing TSAT with its original objective and update rule, following the corresponding paper or official implementation. Table~\ref{tab:first_switch} reports PR, NPR, and SR before and after the first modality switch and over complete sequences. Before includes all frames from the beginning of each sequence up to, but excluding, the first modality-switch frame. After includes the first modality-switch frame and all subsequent
frames until the end of the sequence, including any later modality switches. All includes every frame in the complete sequence. Before switching, methods perform similarly; afterward, their performance diverges, and several general TTA methods fall below the no-adaptation baseline, showing that unreliable updates can impair tracking.

In contrast, TSAT achieves PR, NPR, and SR scores of 46.31\%, 40.36\%, and 34.69\% after modality switching, respectively, ranking first across all three metrics. Compared with the baseline, TSAT improves PR, NPR, and SR by 3.01, 2.86, and 2.49 percentage points, respectively. Compared with the strongest competing result for each metric, TSAT achieves improvements of 2.42, 2.48, and 1.97 percentage points in PR, NPR, and SR, respectively. Moreover, TSAT achieves the best performance across all three metrics over the complete test sequences. These results show that TSAT's spatial-temporal signals and selective updates improve adaptation to tracking-state changes after modality switching.

To further quantitatively evaluate the adaptability of different methods to varying degrees of spatial displacement, we divide the displacement between the ground-truth target centers in two consecutive frames into five ranges: 20--40, 40--60, 60--80, 80--100, and greater than 100 pixels. For events falling within these displacement ranges, we retain only those with non-overlapping evaluation windows and evaluate tracking performance over the subsequent 30 frames of each retained event. Detailed comparisons are reported in Table~\ref{tab:spatial_shift}.

As shown in Table~\ref{tab:spatial_shift}, the performance of most methods generally declines as the spatial displacement increases. In the 20--40, 40--60, and 60--80 pixel ranges, TSAT achieves the best performance across all three metrics, including PR, NPR, and SR. In the 80--100 pixel range, although CoTTA achieves the best results across all three metrics, TSAT ranks second in NPR and SR with scores of 44.43\% and 34.29\%, respectively, and third in PR with a score of 45.76\%. Moreover, TSAT improves PR, NPR, and SR over the SARLA baseline by 3.55, 4.35, and 3.02 percentage points, respectively, demonstrating its effectiveness in improving baseline performance within this displacement range. Under the most severe displacement condition, with displacements exceeding 100 pixels, TSAT maintains the leading performance, achieving 35.39\% PR, 32.51\% NPR, and 27.41\% SR. These results demonstrate that TSAT remains robust under substantial spatial displacement and consistently delivers either leading or competitive performance across the evaluated ranges.

Complementing the short-term evaluation above,
Fig.~\ref{fig:displacement_pr} compares tracking performance
over all remaining frames after spatial displacement.
For each displacement range, we compute PR from its first
occurrence in each eligible sequence to the sequence end,
including the displacement frame, and average the PR scores
across sequences. TSAT outperforms SARLA across all five
ranges, with PR improvements of 4.03--7.63 percentage points,
demonstrating its effectiveness in improving subsequent
tracking performance after spatial displacement.

\subsubsection{Attribute-Based Performance}

To further evaluate the robustness of TSAT under challenging conditions, we compare it with the SARLA baseline~\cite{xiao2026sarla} and six representative trackers, including OSTrack~\cite{ye2022ostrack}, EVPTrack~\cite{shi2024evptrack}, TransT~\cite{chen2021transt}, AQATrack~\cite{xie2024aqatrack}, SeqTrack~\cite{chen2023seqtrack}, and HiT~\cite{kang2023hit}, on the 19 attribute-specific subsets of CM-UOT. These attributes cover diverse challenges, including occlusion, illumination variation, background clutter, UAV and target motion, rotation, low resolution, scale and aspect-ratio changes, and out-of-view conditions. The results are shown in Fig.~\ref{fig:attribute_radar_trackers}.

TSAT achieves the strongest overall attribute-level performance across PR, NPR, and SR, outperforming the compared trackers on most challenge attributes. In particular, TSAT shows notable gains under low resolution
and consistent improvements on horizontal movement, vertical movement, scale variation, and aspect ratio change. Horizontal and vertical movement are closely associated with spatial displacement caused by UAV or target motion, while low resolution, scale variation, and aspect ratio change introduce prominent appearance and geometric variations. Overall, the attribute-based results demonstrate that TSAT remains robust under both
motion-related and appearance-related challenges.

To further evaluate the localization robustness of TSAT under modality switching and spatial variations, Fig.~\ref{fig:tta_qualitative} presents qualitative comparisons among different test-time adaptation methods on CM-UOT.

Before modality switching, most methods can localize the target relatively accurately. After switching, the changes in modality appearance and spatial position make target localization more challenging. In particular, when considerable inter-frame displacement occurs, several methods drift toward nearby similar objects or background regions, resulting in target confusion or bounding-box displacement. In contrast, the predictions of TSAT remain closer to the ground-truth target locations in the presented examples, demonstrating its more stable localization performance under modality switching and spatial variations.

\begin{table}[t]
\centering
\small
\caption{Component ablation of TSAT on CM-UOT.}
\label{tab:ablation}
\renewcommand{\arraystretch}{1.12}
\setlength{\tabcolsep}{4.5pt}
\begin{tabular}{ccccccc}
\toprule
\textbf{No.}
& $\mathcal{L}_{\mathrm{TSE}}$
& $\mathcal{L}_{\mathrm{TBP}}$
& \textbf{RUG}
& \textbf{PR}
& \textbf{NPR}
& \textbf{SR} \\
\midrule
1 &  &  &  & 50.60 & 44.80 & 38.00 \\
2 & \checkmark &  &  & 51.80 & 45.96 & 39.13 \\
3 &  & \checkmark &  & 51.37 & 45.77 & 38.92 \\
4 & \checkmark & \checkmark &  & 52.10 & 46.23 & 39.37 \\
5 & \checkmark &  & \checkmark & 52.78 & 46.92 & 39.91 \\
6 &  & \checkmark & \checkmark & 51.39 & 45.76 & 38.94 \\
7 & \checkmark & \checkmark & \checkmark
  & \textbf{53.19} & \textbf{47.28} & \textbf{40.23} \\
\bottomrule
\end{tabular}
\end{table} 

\subsection{Ablation Study}

\subsubsection{Component Ablation}
To evaluate the effectiveness of the components in TSAT, we conduct component ablation experiments on the CM-UOT dataset, as reported in Table~\ref{tab:ablation}. The baseline achieves PR, NPR, and SR scores of 50.60\%, 44.80\%, and 38.00\%, respectively. Either $\mathcal{L}_{\mathrm{TSE}}$ or $\mathcal{L}_{\mathrm{TBP}}$ improves all metrics, while their combination reaches 52.10\%, 46.23\%, and 39.37\%. Adding RUG achieves 53.19\% PR, 47.28\% NPR, and 40.23\% SR, improving over the baseline by 2.59, 2.48, and 2.23 points. These results validate the components' effectiveness and complementarity.

\begin{table}[t]
\centering
\small
\caption{Center localization error with and without TBP after modality switching (pixels).}
\label{tab:tbp_post_switch}
\renewcommand{\arraystretch}{1.12}
\setlength{\tabcolsep}{5pt}
\begin{tabular}{cccc}
\toprule
\shortstack{Frames since\\switch}
& \shortstack{Without\\TBP}
& \shortstack{With\\TBP}
& \shortstack{Error\\reduction} \\
\midrule
0--9      & 94.75  & 91.89  & 2.86 \\
10--29    & 98.81  & 95.76  & 3.05 \\
30--59    & 106.49 & 102.58 & 3.91 \\
60--119   & 118.55 & 113.30 & 5.25 \\
$\geq120$ & 147.72 & 144.20 & 3.52 \\
\bottomrule
\end{tabular}
\end{table}

\subsubsection{Offline and Online TSE}

To distinguish the effect of TSE as an equivariance constraint during offline
training from its role in test-time adaptation, we compare three settings:
Baseline, Offline TSE, and Online TSE. Offline TSE incorporates the TSE loss
as an auxiliary constraint into the original offline tracking training
objective, while keeping the model parameters fixed during testing. In
contrast, Online TSE constructs the TSE loss from the current unlabeled test
sequence and updates the model parameters online during testing.

As shown in Fig.~\ref{fig:tse_stage_comparison}, Offline TSE achieves PR, NPR,
and SR scores of 48.95\%, 43.53\%, and 37.07\%, respectively. Online TSE
achieves 51.80\%, 45.96\%, and 39.13\%, improving these three metrics by 2.85,
2.43, and 2.06 percentage points, respectively. These results show that,
under the evaluated settings, applying TSE during offline training cannot
replace online adaptation at test time, supporting the effectiveness of using
TSE to adapt the model to the current test sequence.

\begin{table}[t]
\centering
\small
\caption{Effect of the number of Token Shifts $N$ on CM-UOT.}
\label{tab:num_token_shifts}
\renewcommand{\arraystretch}{1.12}
\setlength{\tabcolsep}{8pt}
\begin{tabular}{cccc}
\toprule
$N$ & \textbf{PR (\%)} & \textbf{NPR (\%)} & \textbf{SR (\%)} \\
\midrule
1 & 51.92 & 46.14 & 39.26 \\
2 & 51.76 & 46.01 & 39.14 \\
3 & \textbf{53.19} & \textbf{47.28} & \textbf{40.23} \\
4 & 52.33 & 46.39 & 39.55 \\
5 & 51.62 & 45.86 & 39.04 \\
\bottomrule
\end{tabular}
\end{table}

\subsubsection{Contribution of TBP}
To further examine the contribution of TBP to localization after
modality switching, we compare tracking with and without TBP while
keeping all other settings unchanged. We divide valid visible frames
into five temporal intervals measured from the most recent modality
switch. For each interval, we compute the mean center localization
error for each sequence and then average the results across the
contributing sequences. As shown in Table~\ref{tab:tbp_post_switch},
enabling TBP reduces the mean center error in all five intervals.
In the 60--119-frame interval, the error decreases from 118.55 to
113.30 pixels, a reduction of 5.25 pixels. These results show that
TBP helps reduce the mean center localization error after modality
switching.

\begin{table}[t]
    \centering
    \caption{Ablation on the response entropy threshold $\tau_e$.}
    \label{tab:entropy_threshold}
    \small
    \setlength{\tabcolsep}{9pt}
    \renewcommand{\arraystretch}{1.08}
    \begin{tabular}{c|ccc}
        \toprule
        Threshold $\tau_e$ & PR (\%) & NPR (\%) & SR (\%) \\
        \midrule
        0.16 & 52.00 & 46.17 & 39.28 \\
        0.20 & 52.25 & 46.46 & 39.55 \\
        \textbf{0.24} & \textbf{53.19} & \textbf{47.28} & \textbf{40.23} \\
        0.28 & 52.53 & 46.69 & 39.70 \\
        0.36 & 52.21 & 46.56 & 39.52 \\
        \bottomrule
    \end{tabular}
\end{table}

\begin{table}[t]
    \centering
    \caption{Ablation on the test-time adaptation interval $I$.}
    \label{tab:tta_interval}
    \small
    \setlength{\tabcolsep}{9pt}
    \renewcommand{\arraystretch}{1.08}
    \begin{tabular}{c|ccc}
        \toprule
        Interval $I$ & PR (\%) & NPR (\%) & SR (\%) \\
        \midrule
        5  & 51.09 & 45.41 & 38.69 \\
        8  & 51.45 & 45.87 & 38.90 \\
        \textbf{10} & \textbf{53.19} & \textbf{47.28} & \textbf{40.23} \\
        15 & 52.06 & 46.40 & 39.46 \\
        20 & 51.34 & 45.67 & 38.85 \\
        \bottomrule
    \end{tabular}
\end{table}

\begin{table}[t]
    \centering
    \caption{Ablation on the test-time learning rate $\eta$.}
    \label{tab:tta_learning_rate}
    \small
    \setlength{\tabcolsep}{7pt}
    \renewcommand{\arraystretch}{1.08}
    \begin{tabular}{c|ccc}
        \toprule
        Learning rate $\eta$ & PR (\%) & NPR (\%) & SR (\%) \\
        \midrule
        $5.0\times10^{-6}$ & 50.56 & 45.12 & 38.36 \\
        $4.5\times10^{-5}$ & 52.06 & 46.23 & 39.32 \\
        $\mathbf{6.0\times10^{-5}}$ &
        \textbf{53.19} & \textbf{47.28} & \textbf{40.23} \\
        $7.0\times10^{-5}$ & 52.50 & 46.40 & 39.56 \\
        $1.0\times10^{-4}$ & 51.72 & 46.09 & 39.14 \\
        \bottomrule
    \end{tabular}
\end{table}

\subsubsection{Parameter Analysis}
We further analyze the effect of the number of Token Shifts $N$ in TSE on tracking performance. As defined in Eq.~\eqref{eq:TSE_loss}, TSE constructs $N$ shifted variants of the current search-region token feature map and constrains the corresponding response-center displacements to be consistent with the corresponding sampled Token Shifts. We vary $N$ from 1 to 5, and the results are reported in Table~\ref{tab:num_token_shifts}.

The results show that $N=3$ provides the best overall tracking performance. With few Token Shifts, translation--response relations cover limited spatial variations and may be sensitive to individual responses. Increasing $N$ to 3 provides more representative translation--response correspondences and yields a more stable self-supervised signal. However, further increasing $N$ does not provide additional gains and may introduce redundant or less reliable shifted responses. Despite RUG, excessive Token Shifts can increase adaptation variability. We therefore set $N=3$, balancing shift diversity and response reliability.

\subsubsection{Hyperparameter Analysis}
We examine the influence of the response entropy threshold $\tau_e$, adaptation interval $I$, and test-time learning rate $\eta$ on CM-UOT. The settings used in the main experiments are highlighted in bold.

\paragraph{Response Entropy Threshold}
The response entropy threshold determines which test samples are eligible for online updates in RUG. As reported in Table~\ref{tab:entropy_threshold}, an overly strict threshold excludes many potentially useful samples, whereas a relaxed threshold allows uncertain responses to participate in adaptation. Setting $\tau_e=0.24$ provides the best balance between sample reliability and adaptation opportunity and is therefore used in all experiments.

\begin{figure*}[t]
    \centering
    \includegraphics[width=\textwidth]{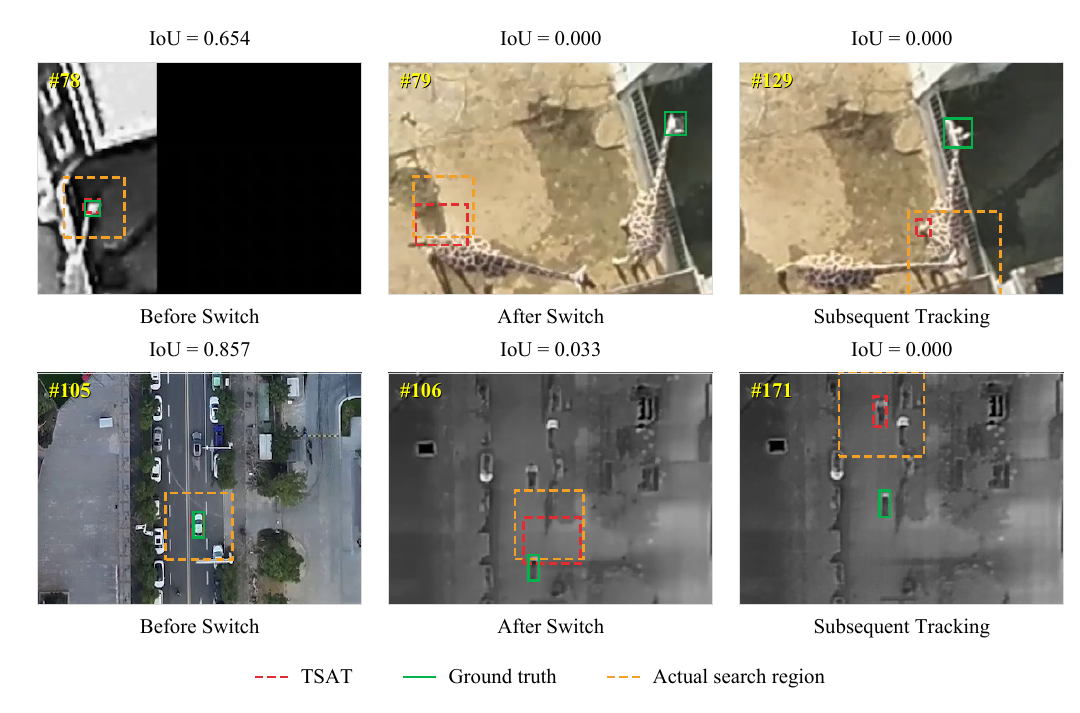}
    \caption{Failure cases of TSAT when the target moves outside the search region after modality switching.}
    \label{fig:failure_cases}
\end{figure*}

\paragraph{Test-Time Adaptation Interval}
The adaptation interval controls how frequently the tracker updates its parameters during inference. As shown in Table~\ref{tab:tta_interval}, short intervals repeatedly optimize the model on highly correlated frames and increase the influence of transient prediction noise, while long intervals delay adaptation to changes in modality and target state. An interval of $I=10$ achieves the best overall performance, balancing timely adaptation with update stability.

\paragraph{Test-Time Learning Rate}
The test-time learning rate determines the magnitude of each online parameter update. Table~\ref{tab:tta_learning_rate} shows that a small learning rate produces only limited adjustment within the available update steps, whereas a large learning rate may amplify noisy frame-level signals and disturb the pretrained representation. The selected learning rate $\eta=6\times10^{-5}$ offers the best trade-off between preserving the pretrained tracker and adapting it to the current test sequence.

\begin{table}[!t]
\centering
\small
\caption{Random-seed analysis of TSAT on CM-UOT.}
\label{tab:random_seed}
\renewcommand{\arraystretch}{1.12}
\setlength{\tabcolsep}{8pt}
\begin{tabular}{ccc}
\toprule
\textbf{PR} & \textbf{NPR} & \textbf{SR} \\
\midrule
53.19 & 47.28 & 40.23 \\
52.97 & 47.09 & 40.12 \\
52.54 & 46.70 & 39.77 \\
52.66 & 46.79 & 39.81 \\
\midrule
$52.84 \pm 0.30$
& $46.97 \pm 0.27$
& $39.98 \pm 0.23$ \\
\bottomrule
\end{tabular}
\end{table}

\subsubsection{Random Seed Analysis}
To assess the stability of TSAT, we repeat the complete evaluation
with four different random seeds while keeping all other
experimental settings unchanged. Table~\ref{tab:random_seed}
reports the individual results, with the mean and sample standard
deviation presented in the last row. The average PR, NPR, and SR
scores are 52.84\%, 46.97\%, and 39.98\%, respectively, with
standard deviations of 0.30, 0.27, and 0.23 percentage points.
These small variations indicate stable tracking performance
across the evaluated random seeds.

\begin{table}[!t]
\centering
\footnotesize
\caption{Efficiency and accuracy on CM-UOT.}
\label{tab:efficiency}
\renewcommand{\arraystretch}{1.15}
\setlength{\tabcolsep}{2.5pt}
\begin{tabular*}{\columnwidth}
{@{\extracolsep{\fill}}lcccccc@{}}
\toprule
\textbf{Method}
& \textbf{FPS}
& \shortstack{\textbf{FLOPs}\\\textbf{(G)}}
& \shortstack{\textbf{Params}\\\textbf{(M)}}
& \textbf{PR}
& \textbf{NPR}
& \textbf{SR} \\
\midrule
Baseline
& \textbf{106.41}
& \textbf{51.1}
& 90.1
& 50.60
& 44.80
& 38.00 \\

TSAT
& 72.87
& 72.2
& 90.1
& \textbf{53.19}
& \textbf{47.28}
& \textbf{40.23} \\
\bottomrule
\end{tabular*}
\end{table}

\subsubsection{Efficiency Analysis}
Table~\ref{tab:efficiency} presents the efficiency and accuracy
comparison between the proposed method and the baseline. The
baseline operates at 106.41 FPS with 51.1 G FLOPs, while our
method achieves 72.87 FPS with 72.2 G FLOPs. The additional
computational cost mainly results from the test-time consistency
constraints and online parameter updates. Nevertheless, our method
still maintains real-time tracking speed without introducing
additional learnable parameters. In terms of tracking accuracy,
our method improves PR, NPR, and SR by 2.59, 2.48, and 2.23
percentage points, respectively. These results demonstrate that
the proposed method achieves a favorable balance between tracking
accuracy and computational efficiency.

\subsection{Generalization Study}

\subsubsection{Evaluation on OSTrack}
To evaluate whether the proposed TSAT can be transferred to a
different tracking framework, we further apply it to OSTrack.
Since TSAT exploits temporal information from the previous frame,
we construct OSTrack$^{\ast}$ by extending the original tracker
to receive three inputs, including the initial template, the
current search region, and the previous-frame search region.
The remaining network components and evaluation settings are
kept unchanged.

As reported in Table~\ref{tab:ostrack_generalization},
OSTrack$^{\ast}$ achieves PR, NPR, and SR scores of 45.45\%,
40.00\%, and 34.09\%, respectively. Incorporating TSAT improves
the corresponding scores to 47.67\%, 42.02\%, and 35.75\%,
resulting in absolute gains of 2.22, 2.02, and 1.66 percentage
points. The consistent improvements across all three evaluation
metrics indicate that TSAT is not restricted to the tracker used
in the main experiments and retains its effectiveness when
transferred to OSTrack.

\begin{table}[!t]
\centering
\small
\caption{Generalization evaluation of TSAT on OSTrack.
All results are reported in percentage.}
\label{tab:ostrack_generalization}

\renewcommand{\arraystretch}{1.08}
\setlength{\tabcolsep}{3pt}

\begin{tabular}{
    p{42pt}
    *{3}{
        >{\centering\arraybackslash}p{24pt}
        @{\hspace{2pt}}
        >{\raggedright\arraybackslash}p{24pt}
    }
}
\toprule
\textbf{Method}
& \textbf{PR} &
& \textbf{NPR} &
& \textbf{SR} & \\
\midrule

OSTrack$^{\ast}$
& 45.45 &
& 40.00 &
& 34.09 & \\

Ours
& \textbf{47.67}
& {\footnotesize\textcolor{blue}{(+2.22)}}
& \textbf{42.02}
& {\footnotesize\textcolor{blue}{(+2.02)}}
& \textbf{35.75}
& {\footnotesize\textcolor{blue}{(+1.66)}} \\

\bottomrule
\end{tabular}

\vspace{2pt}
\begin{minipage}{0.90\columnwidth}
\footnotesize
$^{\ast}$OSTrack is extended to receive the initial template,
the current search region, and the previous-frame search region,
while its remaining components are kept unchanged.
\end{minipage}
\end{table}

\begin{table}[t]
\centering
\small
\caption{Tracking performance on CMOTB.}
\label{tab:cmotb_generalization}

\renewcommand{\arraystretch}{1.08}
\setlength{\tabcolsep}{3pt}

\begin{tabular}{
    p{42pt}
    *{3}{
        >{\centering\arraybackslash}p{24pt}
        @{\hspace{2pt}}
        >{\raggedright\arraybackslash}p{24pt}
    }
}
\toprule
\textbf{Method}
& \textbf{PR} &
& \textbf{NPR} &
& \textbf{SR} & \\
\midrule

OSTrack
& 74.30 &
& 76.80 &
& 63.30 & \\

SARLA
& 76.08 &
& 78.74 &
& 64.85 & \\

TSAT
& \textbf{76.87}
& {\footnotesize\textcolor{blue}{(+0.79)}}
& \textbf{79.46}
& {\footnotesize\textcolor{blue}{(+0.72)}}
& \textbf{65.45}
& {\footnotesize\textcolor{blue}{(+0.60)}} \\

\bottomrule
\end{tabular}
\end{table}

\subsubsection{Evaluation on CMOTB}
To further evaluate the applicability of TSAT to other
cross-modal tracking scenarios, we conduct supplementary
experiments on CMOTB. This dataset contains switches between
RGB and near-infrared (NIR) modalities, enabling evaluation
of localization performance under cross-modal appearance
changes. We use SARLA as the baseline and compare tracking
performance with and without TSAT. The results are presented
in Table~\ref{tab:cmotb_generalization}, where the OSTrack
results are taken from the literature.

On CMOTB, TSAT achieves PR, NPR, and SR scores of
76.87\%, 79.46\%, and 65.45\%, respectively, improving
over SARLA by 0.79, 0.72, and 0.60 percentage points.
For this supplementary experiment, the adaptation
learning rate is set to $5\times10^{-5}$ and the TSE
loss weight to 290. The CM-UOT main-experiment settings
remain unchanged. These results demonstrate that TSAT
further improves the tracking performance of SARLA on
CMOTB, supporting the effectiveness of the proposed
selective online adaptation method on this dataset.

\subsection{Limitations and Future Work}

Despite its improvements under modality switching and spatial displacement, TSAT still has limitations. Its auxiliary tasks operate on the available search-region features and stored tracking states. As illustrated in Fig.~\ref{fig:failure_cases}, a large target-position shift at a modality switch can place the target outside the actual search region. In this situation, the resulting adaptation signals may be insufficient for tracking recovery. Moreover, reliability assessment remains indirect: a similar distractor may produce a concentrated response with low entropy, while an inaccurate historical state may weaken the temporal constraint. Response concentration and branch-loss consistency therefore cannot fully establish target identity.

Future work will explore combining selective adaptation with wider-area search and target re-localization to improve recovery after severe tracking interruptions. Reliability assessment
could also incorporate cross-frame appearance consistency and selectively maintained long-term
references to distinguish stable target responses from confident distractor responses. Furthermore, constructing auxiliary responses and backpropagating the losses introduce additional computational overhead. Investigating update schedules that respond to tracking conditions and more efficient auxiliary-task computation represents a further direction toward deploying TSAT on
resource-constrained UAV platforms.

\section{Conclusion}

In this paper, we propose TSAT, a test-time selective adaptation framework for cross-modal UAV object tracking that introduces two complementary auxiliary tasks, Token Shift Equivariance (TSE) and Temporal Backward Prediction (TBP), to construct localization-oriented spatial and temporal signals for online adaptation. TSE models the correspondence between token translations and localization offsets, while TBP exploits adjacent-frame temporal dependencies through backward prediction. RUG filters unreliable responses and abnormal adaptation signals to stabilize online updates. Experiments on CM-UOT show that TSAT outperforms representative trackers and TTA methods under modality switching and spatial displacement. Ablations validate the proposed components' effectiveness and complementarity. Experiments across tracking frameworks and datasets further show that TSAT's performance gains extend beyond the main experimental setting, supporting its generalization capability.

\bibliographystyle{IEEEtran}
\bibliography{references}

@inproceedings{bertinetto2016siamfc,
  author    = {L. Bertinetto and J. Valmadre and J. F. Henriques and A. Vedaldi and P. H. S. Torr},
  title     = {Fully-Convolutional Siamese Networks for Object Tracking},
  booktitle = {European Conference on Computer Vision Workshops},
  pages     = {850--865},
  year      = {2016}
}

@inproceedings{danelljan2019atom,
  author    = {M. Danelljan and G. Bhat and F. S. Khan and M. Felsberg},
  title     = {{ATOM}: Accurate Tracking by Overlap Maximization},
  booktitle = {IEEE/CVF Conference on Computer Vision and Pattern Recognition},
  pages     = {4660--4669},
  year      = {2019}
}

@inproceedings{bhat2019dimp,
  author    = {G. Bhat and M. Danelljan and L. Van Gool and R. Timofte},
  title     = {Learning Discriminative Model Prediction for Tracking},
  booktitle = {IEEE/CVF International Conference on Computer Vision},
  pages     = {6182--6191},
  year      = {2019}
}

@inproceedings{yan2021stark,
  author    = {B. Yan and H. Peng and J. Fu and D. Wang and H. Lu},
  title     = {Learning Spatio-Temporal Transformer for Visual Tracking},
  booktitle = {IEEE/CVF International Conference on Computer Vision},
  pages     = {10448--10457},
  year      = {2021}
}

@article{li2019rgbtbenchmark,
  author  = {C. Li and X. Liang and Y. Lu and N. Zhao and J. Tang},
  title   = {{RGB-T} Object Tracking: Benchmark and Baseline},
  journal = {Pattern Recognition},
  volume  = {96},
  pages   = {106977},
  year    = {2019}
}

@inproceedings{xiao2022apfnet,
  author    = {Y. Xiao and M. Yang and C. Li and L. Liu and J. Tang},
  title     = {Attribute-Based Progressive Fusion Network for {RGB-T} Tracking},
  booktitle = {AAAI Conference on Artificial Intelligence},
  pages     = {2831--2838},
  year      = {2022}
}

@article{liu2024challenge,
  author  = {L. Liu and C. Li and Y. Xiao and R. Ruan and M. Fan},
  title   = {{RGBT} Tracking via Challenge-Based Appearance Disentanglement and Interaction},
  journal = {IEEE Transactions on Image Processing},
  volume  = {33},
  pages   = {1753--1767},
  year    = {2024}
}

@inproceedings{hui2023tbsi,
  author    = {T. Hui and Z. Xun and F. Peng and J. Huang and X. Wei and X. Wei and J. Dai and J. Han and S. Liu},
  title     = {Bridging Search Region Interaction with Template for {RGB-T} Tracking},
  booktitle = {IEEE/CVF Conference on Computer Vision and Pattern Recognition},
  pages     = {13630--13639},
  year      = {2023}
}

@inproceedings{zhu2023vipt,
  author    = {J. Zhu and S. Lai and X. Chen and D. Wang and H. Lu},
  title     = {Visual Prompt Multi-Modal Tracking},
  booktitle = {IEEE/CVF Conference on Computer Vision and Pattern Recognition},
  pages     = {9516--9526},
  year      = {2023}
}

@inproceedings{cao2024bat,
  author    = {B. Cao and J. Guo and P. Zhu and Q. Hu},
  title     = {Bi-Directional Adapter for Multi-Modal Tracking},
  booktitle = {AAAI Conference on Artificial Intelligence},
  volume    = {38},
  pages     = {927--935},
  year      = {2024}
}

@inproceedings{hou2024sdstrack,
  author    = {X. Hou and J. Xing and Y. Qian and Y. Guo and S. Xin and J. Chen and K. Tang and M. Wang and Z. Jiang and L. Liu and others},
  title     = {{SDSTrack}: Self-Distillation Symmetric Adapter Learning for Multi-Modal Visual Object Tracking},
  booktitle = {IEEE/CVF Conference on Computer Vision and Pattern Recognition},
  pages     = {26551--26561},
  year      = {2024}
}

@inproceedings{zhang2022vtuav,
  author    = {P. Zhang and J. Zhao and D. Wang and H. Lu and X. Ruan},
  title     = {Visible-Thermal {UAV} Tracking: A Large-Scale Benchmark and New Baseline},
  booktitle = {IEEE/CVF Conference on Computer Vision and Pattern Recognition},
  year      = {2022}
}

@article{xiao2026sarla,
  author  = {Yun Xiao and Zhihong Hong and Jiandong Jin and Chenglong Li and
  Jin Tang and Amir Hussain},
  title   = {Cross-Modal UAV Object Tracking: State-Aware Representation Learning and a Unified Benchmark},
  journal = {IEEE Transactions on Circuits and Systems for Video Technology},
  year    = {2026}
}

@inproceedings{li2022crossmodal,
  author    = {C. Li and T. Zhu and L. Liu and X. Si and Z. Fan and S. Zhai},
  title     = {Cross-Modal Object Tracking: Modality-Aware Representations and a Unified Benchmark},
  booktitle = {AAAI Conference on Artificial Intelligence},
  volume    = {36},
  number    = {2},
  pages     = {1289--1296},
  year      = {2022}
}

@article{liu2024modalityfusion,
  author  = {L. Liu and M. Zhang and C. Li and C. Li and J. Tang},
  title   = {Cross-Modal Object Tracking via Modality-Aware Fusion Network and a Large-Scale Dataset},
  journal = {IEEE Transactions on Neural Networks and Learning Systems},
  volume  = {36},
  number  = {4},
  pages   = {6981--6994},
  year    = {2025}
}

@inproceedings{li2024multiteacher,
  author    = {Y. Li and L. Liu and M. Zhang and C. Li},
  title     = {Multi-Teacher Knowledge Distillation with Triplet Loss for Cross-Modal Object Tracking},
  booktitle = {International Conference on Brain Inspired Cognitive Systems},
  pages     = {155--165},
  year      = {2024}
}

@article{liu2025prototype,
  author  = {L. Liu and C. Li and F. Wang and L. Shen and J. Tang},
  title   = {Prototype-Based Cross-Modal Object Tracking},
  journal = {Information Fusion},
  volume  = {118},
  pages   = {102941},
  year    = {2025}
}

@inproceedings{sun2020ttt,
  author    = {Y. Sun and X. Wang and Z. Liu and J. Miller and A. A. Efros and M. Hardt},
  title     = {Test-Time Training with Self-Supervision for Generalization under Distribution Shifts},
  booktitle = {International Conference on Machine Learning},
  pages     = {9229--9248},
  year      = {2020}
}

@inproceedings{wang2021tent,
  author    = {D. Wang and E. Shelhamer and S. Liu and B. Olshausen and T. Darrell},
  title     = {{TENT}: Fully Test-Time Adaptation by Entropy Minimization},
  booktitle = {International Conference on Learning Representations},
  year      = {2021}
}

@inproceedings{wang2022cotta,
  author    = {Q. Wang and O. Fink and L. Van Gool and D. Dai},
  title     = {Continual Test-Time Domain Adaptation},
  booktitle = {IEEE/CVF Conference on Computer Vision and Pattern Recognition},
  pages     = {7201--7211},
  year      = {2022}
}

@inproceedings{niu2022eata,
  author    = {S. Niu and J. Wu and Y. Zhang and Y. Chen and S. Zheng and P. Zhao and M. Tan},
  title     = {Efficient Test-Time Model Adaptation without Forgetting},
  booktitle = {International Conference on Machine Learning},
  pages     = {16888--16905},
  year      = {2022}
}

@inproceedings{niu2023sar,
  author    = {S. Niu and J. Wu and Y. Zhang and Z. Wen and Y. Chen and P. Zhao and M. Tan},
  title     = {Towards Stable Test-Time Adaptation in Dynamic Wild World},
  booktitle = {International Conference on Learning Representations},
  year      = {2023}
}

@inproceedings{shao2025pura,
  author    = {Z. Shao and Y. Hu and B. Fan and H. Liu},
  title     = {{PURA}: Parameter Update-Recovery Test-Time Adaption for {RGB-T} Tracking},
  booktitle = {IEEE/CVF Conference on Computer Vision and Pattern Recognition},
  pages     = {22089--22098},
  year      = {2025}
}

@inproceedings{dosovitskiy2021vit,
  author    = {A. Dosovitskiy and others},
  title     = {An Image Is Worth 16$\times$16 Words: Transformers for Image Recognition at Scale},
  booktitle = {International Conference on Learning Representations},
  year      = {2021}
}

@article{ba2016layernorm,
  author  = {J. L. Ba and J. R. Kiros and G. E. Hinton},
  title   = {Layer Normalization},
  journal = {arXiv preprint arXiv:1607.06450},
  year    = {2016}
}

@inproceedings{chen2021transt,
  author    = {X. Chen and B. Yan and J. Zhu and D. Wang and X. Yang and H. Lu},
  title     = {Transformer Tracking},
  booktitle = {IEEE/CVF Conference on Computer Vision and Pattern Recognition},
  pages     = {8126--8135},
  year      = {2021}
}

@inproceedings{ye2022ostrack,
  author    = {B. Ye and H. Chang and B. Ma and S. Shan and X. Chen},
  title     = {Joint Feature Learning and Relation Modeling for Tracking: A One-Stream Framework},
  booktitle = {European Conference on Computer Vision},
  pages     = {341--357},
  year      = {2022}
}

@inproceedings{kang2023hit,
  author    = {B. Kang and X. Chen and D. Wang and H. Peng and H. Lu},
  title     = {Exploring Lightweight Hierarchical Vision Transformers for Efficient Visual Tracking},
  booktitle = {IEEE/CVF International Conference on Computer Vision},
  pages     = {9612--9621},
  year      = {2023}
}

@inproceedings{chen2023seqtrack,
  author    = {X. Chen and H. Peng and D. Wang and H. Lu and H. Hu},
  title     = {{SeqTrack}: Sequence to Sequence Learning for Visual Object Tracking},
  booktitle = {IEEE/CVF Conference on Computer Vision and Pattern Recognition},
  pages     = {14572--14581},
  year      = {2023}
}

@inproceedings{gao2023grm,
  author    = {S. Gao and C. Zhou and J. Zhang},
  title     = {Generalized Relation Modeling for Transformer Tracking},
  booktitle = {IEEE/CVF Conference on Computer Vision and Pattern Recognition},
  pages     = {18686--18695},
  year      = {2023}
}

@inproceedings{shi2024evptrack,
  author    = {L. Shi and B. Zhong and Q. Liang and N. Li and S. Zhang and X. Li},
  title     = {Explicit Visual Prompts for Visual Object Tracking},
  booktitle = {AAAI Conference on Artificial Intelligence},
  volume    = {38},
  number    = {5},
  pages     = {4838--4846},
  year      = {2024}
}

@inproceedings{li2024avtrack,
  author    = {Y. Li and M. Liu and Y. Wu and X. Wang and X. Yang and S. Li},
  title     = {Learning Adaptive and View-Invariant Vision Transformer for Real-Time {UAV} Tracking},
  booktitle = {International Conference on Machine Learning},
  pages     = {28403--28420},
  year      = {2024}
}

@inproceedings{xie2024aqatrack,
  author    = {J. Xie and B. Zhong and Z. Mo and S. Zhang and L. Shi and S. Song and R. Ji},
  title     = {Autoregressive Queries for Adaptive Tracking with Spatio-Temporal Transformers},
  booktitle = {IEEE/CVF Conference on Computer Vision and Pattern Recognition},
  pages     = {19300--19309},
  year      = {2024}
}

@inproceedings{chen2025sutrack,
  author    = {X. Chen and B. Kang and W. Geng and J. Zhu and Y. Liu and D. Wang and H. Lu},
  title     = {{SUTrack}: Towards Simple and Unified Single Object Tracking},
  booktitle = {AAAI Conference on Artificial Intelligence},
  volume    = {39},
  number    = {2},
  pages     = {2239--2247},
  year      = {2025}
}

@inproceedings{zhang2026spiketrack,
  author    = {Qiuyang Zhang and Jiujun Cheng and Qichao Mao and Cong Liu and
  Yu Fang and Yuhong Li and Mengying Ge and Shangce Gao},
  title     = {SpikeTrack: A Spike-driven Framework for Efficient Visual Tracking},
  booktitle = {Proceedings of the IEEE/CVF Conference on Computer Vision and Pattern Recognition},
  pages     = {6802--6811},
  year      = {2026}
}

@article{li2018adabn,
  author  = {Y. Li and N. Wang and J. Shi and X. Hou and J. Liu},
  title   = {Adaptive Batch Normalization for Practical Domain Adaptation},
  journal = {Pattern Recognition},
  volume  = {80},
  pages   = {109--117},
  year    = {2018}
}

@inproceedings{zhang2022memo,
  author    = {M. Zhang and S. Levine and C. Finn},
  title     = {{MEMO}: Test Time Robustness via Adaptation and Augmentation},
  booktitle = {Advances in Neural Information Processing Systems},
  volume    = {35},
  pages     = {38629--38642},
  year      = {2022}
}

@inproceedings{yuan2023rotta,
  author    = {L. Yuan and B. Xie and S. Li},
  title     = {Robust Test-Time Adaptation in Dynamic Scenarios},
  booktitle = {IEEE/CVF Conference on Computer Vision and Pattern Recognition},
  pages     = {15922--15932},
  year      = {2023}
}

@inproceedings{lee2024deyo,
  author    = {J. Lee and D. Jung and S. Lee and J. Park and J. Shin and U. Hwang and S. Yoon},
  title     = {Entropy Is Not Enough for Test-Time Adaptation: From the Perspective of Disentangled Factors},
  booktitle = {International Conference on Learning Representations},
  year      = {2024}
}

@inproceedings{zaveri2025siamabc,
  author    = {Ram Zaveri and Shivang Patel and Yu Gu and Gianfranco Doretto},
  title     = {Improving Accuracy and Generalization for Efficient Visual Tracking},
  booktitle = {IEEE/CVF Winter Conference on Applications of Computer Vision},
  pages     = {9468--9478},
  year      = {2025}
}

@inproceedings{lai2026gold,
  author    = {Guannan Lai and Da-Wei Zhou and Zhenguo Li and Han-Jia Ye},
  title     = {The Golden Subspace: Where Efficiency Meets Generalization in Continual Test-Time Adaptation},
  booktitle = {IEEE/CVF Conference on Computer Vision and Pattern Recognition},
  pages     = {3866--3875},
  year      = {2026}
}

@inproceedings{baek2026imse,
  author    = {Sunghyun Baek and Jaemyung Yu and Seunghee Koh and Minsu Kim and Hyeonseong Jeon and Junmo Kim},
  title     = {{IMSE}: Intrinsic Mixture of Spectral Experts Fine-tuning for Test-Time Adaptation},
  booktitle = {International Conference on Learning Representations},
  year      = {2026}
}

@inproceedings{zhang2026curvature,
  title     = {Curvature-Aware Zeroth-Order Optimization for Memory-Efficient Test-Time Adaptation},
  author    = {Zhang, Junming and Yin, Shuyu and Liu, Peilin and Ying, Rendong and Wen, Fei},
  booktitle = {Proceedings of the IEEE/CVF Conference on Computer Vision and Pattern Recognition},
  pages     = {836--846},
  year      = {2026}
}

@inproceedings{segu2023darth,
  author    = {Segu, Mattia and Schiele, Bernt and Yu, Fisher},
  title     = {{DARTH}: Holistic Test-time Adaptation for
               Multiple Object Tracking},
  booktitle = {Proceedings of the IEEE/CVF International
               Conference on Computer Vision},
  pages     = {9717--9727},
  year      = {2023}
}

@inproceedings{bertrand2023testtime,
  author    = {Bertrand, Juliette and Kordopatis Zilos, Giorgos
               and Kalantidis, Yannis and Tolias, Giorgos},
  title     = {Test-time Training for Matching-based
               Video Object Segmentation},
  booktitle = {Advances in Neural Information Processing Systems},
  volume    = {36},
  pages     = {20918--20941},
  year      = {2023}
}

\begin{IEEEbiographynophoto}{Yun Xiao}
received the M.S. and Ph.D. degrees from the School of
Computer Science and Technology, Anhui University,
Hefei, China, in 2011 and 2019, respectively.
She is currently an Associate Professor with the School
of Artificial Intelligence, Anhui University.
Her research interests include computer vision and
deep learning.
\end{IEEEbiographynophoto}

\begin{IEEEbiographynophoto}{Hailong Ye}
is currently pursuing a master's degree at Anhui
University, Hefei, China.
His research interests include computer vision,
cross-modal object tracking, and test-time adaptation.
\end{IEEEbiographynophoto}

\begin{IEEEbiographynophoto}{Zhicheng Zhao}
is currently an Associate Professor with the School
of Artificial Intelligence, Anhui University,
Hefei, China.
His research interests include computer vision,
multimodal learning, and remote sensing image analysis.
\end{IEEEbiographynophoto}

\begin{IEEEbiographynophoto}{Lei Liu}
received the B.S. and Ph.D. degrees from Anhui
University, Hefei, China, in 2019 and 2024, respectively.
He is currently with the School of Artificial
Intelligence, Anhui University of Science and Technology,
Huainan, China.
His research interests include computer vision and
deep learning.
\end{IEEEbiographynophoto}

\begin{IEEEbiographynophoto}{Amir Hussain}
received the B.Eng. (Hons.) and Ph.D. degrees in
electronic and electrical engineering from the
University of Strathclyde, Glasgow, U.K., in 1992
and 1997, respectively.
He is currently a Chair Professor with the Department
of Information and Computer Science, King Fahd
University of Petroleum and Minerals, Dhahran,
Saudi Arabia.
His research interests include artificial intelligence,
cognitive data science, and their applications
in healthcare and industrial systems.
\end{IEEEbiographynophoto}
\vfill

\end{document}